\documentclass[preprint,12pt]{elsarticle}

\usepackage[utf8]{inputenc}
\usepackage[T1]{fontenc}
\usepackage{amsmath, amssymb, amsthm}
\usepackage{graphicx}
\usepackage{booktabs}
\usepackage[colorlinks=true,linkcolor=black,citecolor=black,urlcolor=black,
            pdfauthor={Xue Qin and Simin Luan and Cong Yang and Zhijun Li},
            pdftitle={Evolving Skill Modules under a Fixed Planner: Versioning, Rollback, and Runtime Governance for Long-Lived Robot Systems}]{hyperref}
\usepackage{cleveref}
\usepackage{enumitem}
\usepackage{xcolor}
\usepackage[font=normalsize,labelfont=bf]{caption}

\usepackage{placeins}  
\usepackage[protrusion=true,expansion=true,final]{microtype}
\usepackage{tikz}
\usetikzlibrary{arrows.meta, positioning, shapes.geometric, calc, decorations.pathreplacing, backgrounds}
\usepackage{pgfplots}
\pgfplotsset{compat=1.17}
\usepgfplotslibrary{fillbetween}

\newtheorem{definition}{Definition}

\journal{Journal of Systems and Software}

\begin{document}

\begin{frontmatter}

\title{Evolving Skill Modules under a Fixed Planner: Versioning, Rollback, and Runtime Governance for Long-Lived Robot Systems}

\author[hitsw]{Xue Qin}
\ead{qinxue@me.com}
\author[hitcs]{Simin Luan}
\ead{luansiminiot@gmail.com}
\author[sudafse]{Cong Yang\corref{cor1}}
\ead{cong.yang@suda.edu.cn}
\author[hitcs]{Zhijun Li\corref{cor1}}
\ead{lizhijun_os@hit.edu.cn}
\cortext[cor1]{Corresponding authors.}

\affiliation[hitsw]{organization={School of Software, Harbin Institute of Technology}, city={Harbin}, country={China}}
\affiliation[hitcs]{organization={School of Computer Science and Technology, Harbin Institute of Technology}, city={Harbin}, country={China}}
\affiliation[sudafse]{organization={School of Future Science and Engineering, Soochow University}, city={Suzhou}, country={China}}

\begin{abstract}
Robots deployed for long periods keep improving their skills, and each update changes a released system. We treat this as a software-lifecycle problem: a fixed decision layer dispatches versioned skill modules and a runtime layer was built to screen each action. On six robosuite tasks we report three negative results and two measurements. First, peak task success is unstable across random seeds (within one method it spans $23.3$ to $73.3\%$), so single-run peaks cannot rank these methods. Second, the system's four modules are whole-task policies with different labels, rotated on a clock, not the phase decomposition its documentation describes. At a matched budget one such policy holds the geometry at the final step in $0.734$ of episodes reaching it, averaged over seeds, against $0.023$ for the rotation, with no seed overlap at four seeds per arm (exact $p{=}0.029$). An intervention isolates why: restoring the termination condition the clock replaced raises retention on every seed. Third, our shield cut violations $98$ to $100\%$ on five single-arm tasks ($34.9\%$ on the sixth) by discarding whole actions, leaving success at zero: its acceptance criterion omitted completions, so a shield that stopped the robot scored perfectly. What survives is release machinery: a promotion gate kept all twelve injected regressions out, a rate its calibration nearly guarantees, at a $22.5\%$ clean-candidate rejection cost; a dip detector caught nine of twelve, missing all three on one seed.
\end{abstract}


\end{frontmatter}

\begin{figure*}[!t]
  \centering
  \includegraphics[width=0.65\textwidth]{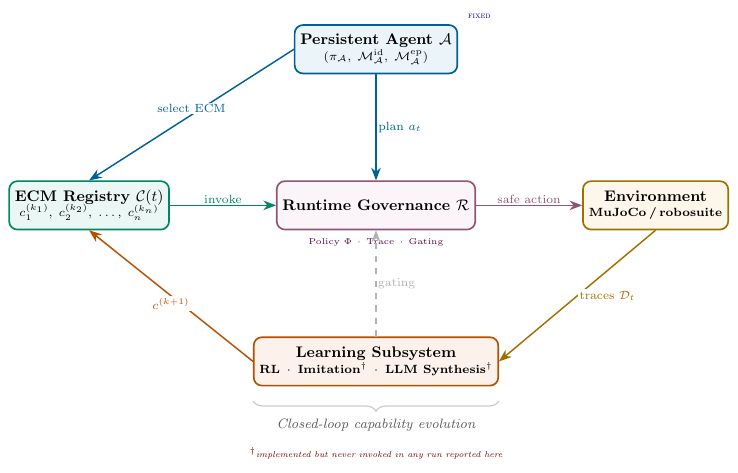}
  \caption{Capability evolution loop for embodied agents. A fixed decision layer (blue, top) selects and sequences modules, while the capability modules (ECMs) themselves evolve over time through iterative execution, experience collection, and learning. A runtime layer (purple) screens each proposed action against configured predicates; \S\ref{sec:exp3} reports the measured behaviour of this layer, including the failure mode that rules out a safety claim for it. The learning subsystem admits three modalities by design, but the imitation and code-synthesis paths (marked $\dagger$) never executed in any run reported in this paper, and reinforcement learning is the only modality behind every number below (\S\ref{sec:learning}).}
  \label{fig:evolution-loop}
\end{figure*}

\section{Introduction}
\label{sec:introduction}

Embodied agents are expected to operate persistently in the physical world, interacting with dynamic environments over extended periods of time. Unlike short-lived software agents, such systems must continuously acquire new capabilities while maintaining stable behavior and consistent control.

\looseness=-1 Improvement after release is what makes such systems hard to maintain. A robot that keeps learning is a deployed software system whose components are being replaced while it runs, and the replacement is opaque in a way ordinary software updates are not: a retrained skill carries no changelog, its behaviour is checked statistically rather than by unit tests, and a regression may surface only in the field. Two distinct kinds of change are often conflated here. Editing an agent's prompt, tools, or episodic context leaves the underlying decision parameters untouched, whereas retraining or fine-tuning changes them; only the second makes the deployed decision function itself a moving target, and it is the second that undermines the ability to say which change caused which behaviour.

This matters more in embodied settings, where instability in decision-making leads to unsafe or irreversible physical outcomes rather than a bad answer. A deployed robot therefore needs what deployed software has long had: explicit versions for the parts that change, a gate that decides whether a new version may be promoted, a way back when it should not have been, and an enforcement boundary that holds regardless of which version is live.

Existing work supplies parts of this and not the whole. Hierarchical control decomposes behaviour into reusable sub-policies through options~\citep{sutton1999options}, option-critic~\citep{bacon2017option}, and HIRO~\citep{nachum2018hiro}, and skill-library methods such as SPiRL~\citep{pertsch2021spirl}, SkiMo~\citep{shi2023skimo}, and Voyager~\citep{wang2023voyager} accumulate reusable behaviours; safe RL constrains or shields a policy during learning~\citep{garcia2015comprehensive}, including per-action safety layers that project each action into a safe set~\citep{dalal2018safelayer}; continual learning targets parameter-level forgetting~\citep{khetarpal2022continual,parisi2019continual}. Systems built specifically for evolving embodied agents have appeared while this work was under way, and they set the current bar: PhyAgentOS~\citep{liu2026phyagentos} runs a full operating system on real robots, and CaP-X~\citep{fu2026capx} benchmarks coding agents across twelve frontier models and hundreds of manipulation tasks. Mainstream MLOps does provide lifecycles for learned models: TFX ships validation gates, versioned artifacts, and safe rollback in production~\citep{baylor2017tfx}, ease.ml/ci treats model promotion as statistically rigorous continuous integration~\citep{renggli2019easemlci}, and the surveyed MLOps pipelines standardise versioning and rollback around retraining~\citep{kreuzberger2023mlops}. Robot learning on the job places deployment-time retraining behind human oversight that decides how much autonomy each policy version receives~\citep{liu2023sirius}, a promotion decision made by a person. Concurrent work mechanises part of the decision for robot policies: AgenticRobotics admits a checkpoint only on recorded evidence inside an automated train--evaluate--improve loop and measures its false-promotion rate~\citep{yu2026agenticrobotics}. Its gate chooses among checkpoints during development, and no path for reverting a regression after promotion is reported. What we found no executable treatment of is that lifecycle made mechanical, wrapped around independently retrained skill modules inside a robot runtime, with its gate, rollback, and enforcement costs measured on one system; our contribution is that measured case, not the lifecycle concept. In PhyAgentOS and CaP-X the evolving artifact is already legible: PhyAgentOS evolves a memory of verified lessons without retraining, and CaP-X evolves programs that can be read and diffed, so both can be reviewed before they ship and reverted by ordinary means. None of SPiRL, SkiMo, or Voyager admits or refuses a retrained learned module against the one in use, and Voyager's library holds programs, the legible kind of artifact discussed below. In safe RL, constrained optimisation and recovery policies tie the guarantee to the policy being trained, so it does not survive that policy being swapped. Runtime shielding is not tied to one policy, and our governance layer belongs to that family. The software-engineering side of the problem is not unexplored either, and it deserves a sharper statement of what is left rather than a claim of absence. Release engineering has versioning, staged promotion, and rollback~\citep{savor2016continuous}; runtime assurance has the Simplex architecture and its verified envelope around an unverified controller~\citep{sha2001simplex}; self-adaptive systems have the MAPE-K loop and a run-time model of the deployed configuration~\citep{kephart2003autonomic,blair2009models}; and the trouble learned components cause for all of it has been catalogued~\citep{sculley2015debt,amershi2019se4ml,paleyes2022challenges}. What that work assumes, and what a retrained policy denies it, is a component whose behaviour can be specified, tested on a fixed input, and diffed between versions. Our contribution is a measurement of what the same machinery costs once that assumption fails.

We describe a runtime architecture built around that missing lifecycle. A fixed high-level decision layer selects and sequences \emph{Embodied Capability Modules} (ECMs), self-contained units of embodied functionality that are versioned, promoted through a gate, and rolled back when a promotion regresses; every action they emit is meant to pass through a governance layer at the execution boundary, independently of which module produced it or how it was trained; where that layer actually ran, and what it did there, is itself one of this paper's results. That layer is another place where description outran code, and \Cref{sec:exp3} reports what it actually does. Holding the decision layer fixed is a standard design choice in hierarchical control rather than a new one, and we make no novelty claim for it; its role here is to make attribution exact, since with the decision layer invariant the provenance of a behaviour is unambiguous: only a module version can have changed. \Cref{fig:evolution-loop} illustrates the architecture.

Why modules at all, if a single learned component would do? Not for task performance: a budget-matched control reported in \Cref{sec:exp-nophase} refutes the one arrangement we built, in which four separately trained whole-task policies take turns driving a single task, and the claim drawn from that arrangement does not survive the control. The reason that survives is a property of deployment rather than a design preference. A robot in service holds many capabilities at once, and they do not share a schedule: a grasping model is retrained when a new gripper arrives, a navigation stack when a site is remapped, an inspection routine when a defect class is added. Those components have independent provenance, independent retraining cadences, and independent regression risk, whether or not partitioning any single task helps solve it. Release machinery is needed because that plurality exists, and it would be needed for one learned module as much as for twenty, since the admission question is asked of the artifact and not of the architecture.

We evaluate on a six-task robosuite manipulation suite, and report three negative results alongside two measurements. Peak success on the long-horizon task swings by a factor of three across random seeds within a single method (\Cref{sec:exp2}), so we stop using single-run peaks to rank methods. A budget-matched control then shows that spending a fixed interaction budget on one whole-task policy beats spreading it across four whole-task policies that take turns on a fixed schedule, by a wide margin and for a reason we can point to in the code (\Cref{sec:exp-nophase}). The third is about our own instrument rather than about a design. We built an action-level safety shield, scored it by how many violations it removed, and got a near-perfect number while the shield was discarding the policy's entire action on most steps and holding task success at zero on every task (\Cref{sec:exp3}). What we can measure and defend is the release machinery: what a promotion gate catches and what it costs, and what a rollback detector misses.

Our contributions are summarized as follows:
\begin{itemize}[nosep]
    \item A lifecycle for learned skill modules, giving each module an explicit version, a promotion gate that admits a candidate only if it holds up on held-out evaluation, and a rollback path when a promoted version regresses.
    \item A runtime governance layer at the execution boundary, and a negative result about how we evaluated it. Scored by violation count the layer looks near-perfect. Scored by whether the task still completes it is a brake: a keep-out zone configured across the middle of the workspace makes one of its four predicates discard the whole action vector on $75$ to $100\%$ of steps, and success under governance is $0.0\%$ on all six tasks (\Cref{sec:exp3}).
    \item An implementation of the closed loop, in which task execution, experience collection, module updating, and reintegration run end to end on a robot simulator. The replication artifact accompanying this submission contains the evaluation drivers, per-seed result files, analysis and figure scripts, and pinned dependencies, and is published at the repository linked in \Cref{sec:exp-setup}; \Cref{sec:provenance} states exactly which code stands behind which number.
    \item An empirical account of what does and does not rank these methods: single-run peak success swings by a factor of three across seeds within one method and cannot serve as an endpoint, whereas retention, the fraction of episodes that reach the success geometry and still hold it when the episode ends, separates the budget-matched arms completely and at every seed (\Cref{sec:exp-nophase,sec:exp-mechanism}).
\end{itemize}

We structure our evaluation around four hypotheses:
\begin{itemize}[nosep,leftmargin=1.8em,label={}]
    \item \textbf{H1:}~Evolving a set of capability modules outperforms monolithic agent modification in mean peak success on long-horizon tasks, while the agent's decision policy stays fixed. \emph{Status: refuted; the comparison confounds architecture with a tenfold budget, and the budget-matched control reverses it (\Cref{sec:exp-nophase}).}
    \item \textbf{H2:}~Runtime governance ensures safe execution even when evolving capabilities attempt unsafe actions. \emph{Status: refuted; the shield suppresses violations by discarding the action, and task success under governance is zero (\Cref{sec:exp3}).}
    \item \textbf{H3:}~Gated promotion keeps a regressed module version out of deployment, and rollback restores service when one reaches deployment. \emph{Status: supported, with measured costs (\Cref{sec:exp-lifecycle}).}
    \item \textbf{H4:}~The ladder step at which episodes stop separates the configurations. \emph{Status: premise supported, expected reading inverted; retention rather than progress is what distinguishes the arms (\Cref{sec:exp-mechanism}).}
\end{itemize}

The remainder of this paper is organised as follows. \Cref{sec:related} positions the work against hierarchical control and the options framework, skill libraries, safe reinforcement learning, runtime assurance and self-adaptive systems, contemporary self-evolving embodied systems, and deployment practice for learned components. \Cref{sec:method} gives the architecture: the formal model of the decision layer and its capability modules, then the module lifecycle, the runtime governance layer, and the learning modalities. \Cref{sec:experiments} reports the evaluation, \Cref{sec:discussion} discusses what the results do and do not support, and \Cref{sec:conclusion} concludes.

\section{Related Work}
\label{sec:related}

This work draws on three literatures: modular skill learning, deployment practice for software that keeps changing, and runtime safety enforcement. All three are developed, and every mechanism we build is standard in at least one of them. We therefore argue the contribution against that work rather than asserting it. Versioning, staged promotion, and rollback are ordinary continuous-delivery practice; enforcement at the execution boundary is the shielding construct of safe reinforcement learning and the switching construct of the Simplex architecture; keeping a model of the running configuration and acting on it while the system runs is the subject of self-adaptive systems research. The paragraphs below state, for each literature, what it already supplies, so that what is left over is visible and can be judged.

\paragraph{Modular skill learning and libraries}
Hierarchical RL decomposes policies into reusable sub-policies via options~\citep{sutton1999options}, option-critic~\citep{bacon2017option}, and HIRO~\citep{nachum2018hiro}. The options definition is worth restating in full, because \Cref{sec:exp-nophase} reports what happens when part of it is left out. An option is a triple of an initiation set, a policy, and a termination function $\beta$ that decides at each state whether control returns to the caller~\citep{sutton1999options}. The $\beta$ component is part of the definition rather than an implementation choice, and methods that discover options learn it alongside the policy~\citep{bacon2017option}. Skill chaining goes further and constructs each skill so that its initiation set covers the states from which the next skill can take over~\citep{konidaris2009chaining}, an idea carried into deep function approximation by \citet{bagaria2020dsc}. The configuration evaluated in this paper has a policy per module and no $\beta$ of any kind: handover is decided by a step counter, and \Cref{sec:exp-nophase} reports the cost. Modular policy architectures factor behaviour into components that transfer across tasks and robots~\citep{devin2017modular}, and modular lifelong RL composes such components to limit interference across a task sequence~\citep{mendez2022modular}. Learned skill libraries extend this idea: unsupervised skill discovery acquires a behaviour repertoire without task reward~\citep{eysenbach2019diayn}, SPiRL~\citep{pertsch2021spirl} extracts skill priors from offline data, SkiMo~\citep{shi2023skimo} jointly learns skills with a dynamics model, and Voyager~\citep{wang2023voyager} acquires skills through LLM-driven exploration. LLM-based systems such as SayCan~\citep{ahn2022saycan}, Code as Policies~\citep{liang2023code}, and Toolformer~\citep{schick2023toolformer} compose or invoke modular capabilities at inference time. Most recently, \citet{tziafas2024lifelong} bootstrap composable skill libraries using LLMs. Our ECMs share the spirit of modularity but differ in two key respects: (i)~ECMs are learned modules versioned through a lifecycle (creation, deployment, evolution, deprecation) in which a retrained version is admitted or refused against the deployed one, which none of the methods above does for a learned skill; and (ii)~ECM evolution is decoupled from the agent's decision policy, which remains fixed.

\paragraph{Continual learning and agent adaptation}
Continual learning methods such as EWC~\citep{kirkpatrick2017overcoming}, progressive networks~\citep{rusu2016progressive}, and PathNet~\citep{fernando2017pathnet} mitigate forgetting at the parameter level, primarily in classification settings~\citep{parisi2019continual,vandeven2022three}; recent reviews survey the continual-RL setting specifically~\citep{khetarpal2022continual}. PathNet, which uses evolutionary path selection in modular networks, is a direct intellectual precursor to our notion of capability evolution through modular structures. The embodied case is comparatively underexplored: dedicated robotic lifelong-learning benchmarks such as Continual World~\citep{wolczyk2021continual} and LIBERO~\citep{liu2023libero}, together with methods that preserve and recombine skills across a task stream~\citep{meng2025preserving}, have emerged only recently. Our approach sidesteps interference architecturally: because each ECM has independent parameters, updating one capability does not affect others, and the agent's decision policy is never modified, eliminating the primary source of forgetting in agent-centric approaches.

\paragraph{LLM-based autonomous agents}
LLM-based agents such as ReAct~\citep{yao2023react}, Reflexion~\citep{shinn2023reflexion}, and Toolformer~\citep{schick2023toolformer} improve via prompt, memory, or tool-invocation modification, which changes the agent's effective policy across iterations. Our framework instead keeps the LLM-based agent as a persistent cognitive core $\mathcal{A}$ and channels improvement through ECM evolution.

\paragraph{Versioned deployment in software systems}
The machinery we build for learned modules is standard practice for ordinary software, and the differences are what make this paper necessary. Continuous-delivery pipelines version every artifact, promote a candidate only after it passes a gate, and keep a path back when a release misbehaves~\citep{savor2016continuous}; canary and blue-green deployments expose a new version to part of the traffic before committing to it. The surface similarity to an ECM lifecycle is close enough that the vocabulary transfers directly. The mechanisms do not. A microservice encapsulates deterministic logic behind a declared interface, so its gate can be a unit test and its correctness is decidable on a fixed input. An ECM wraps a learned stochastic policy whose behaviour shifts with every training iteration, so admission has to be decided statistically on held-out episodes, and the decision inherits the variance of that estimate: our own gate rejects one clean candidate in five at the threshold we use. \citet{sculley2015debt} catalogue the maintenance burden this creates, in which entanglement and undeclared consumers make the effect of retraining one component hard to bound, and \citet{breck2017mltest} propose a readiness rubric whose tests are statistical rather than functional. Rollback differs in the same way. Reverting a service restores a known-good state, whereas reverting a module restores parameters whose competence depends on an environment that may itself have drifted since that version was trained. Finally, composition in our setting runs over continuous action spaces under safety constraints rather than discrete request-response calls, which is why the enforcement boundary of \Cref{sec:runtime} sits between the module and the actuator rather than at an API surface. The practice we borrow from has itself been studied empirically, and the studies bound what is new here. \citet{amershi2019se4ml} document how machine-learning components disturb established software processes, model management and versioning among them, and \citet{ashmore2021assuring} and \citet{paleyes2022challenges} survey what assurance and deployment demand across such a component's lifecycle. Two pieces of that practice are close enough to ours to name precisely. A model registry gives a trained artifact an identity, a version, and a promotion stage, which is what \Cref{sec:ecm-lifecycle} does for an ECM; the difference is that a registry records a promotion decision taken elsewhere, whereas the gate measured in \Cref{sec:exp-lifecycle} takes one. Automated canary analysis decides statistically whether a newly exposed version behaves worse than the incumbent~\citep{tarvo2015canary,schermann2018live}, which is the same inference our gate performs, run on held-out episodes instead of production traffic and against a reward estimate instead of a service health signal. What we claim is therefore not the gate but its application to a component whose output is a continuous action sequence, together with the rejection and miss rates \Cref{sec:exp-lifecycle} reports for it.

\paragraph{Safe reinforcement learning}
Safe RL incorporates constraints into learning, either by optimising the policy subject to constraints (e.g.\ constrained policy optimisation~\citep{achiam2017constrained} and Lagrangian methods), by shielding actions against a safety specification at runtime~\citep{alshiekh2018safe}, or by learning a recovery policy that steers the system back from near-unsafe states~\citep{thananjeyan2021recovery}; broad surveys cover these families~\citep{garcia2015comprehensive,brunke2022safe}. These mechanisms differ in where they sit: constrained optimisation and recovery policies are tied to the policy being trained, whereas shielding is already an execution-boundary construct and is not inherently coupled to one policy. Our runtime governance is of the latter kind, and we claim no novelty for the idea. The design intent was to pair such a boundary with a module lifecycle, so that the envelope would be invariant across module versions rather than tied to whichever version is live. What we can report is not that pairing working but the way we mismeasured it: a shield of this shape, evaluated on violations alone, earns a near-perfect score by stopping the robot (\Cref{sec:exp3}). The shielding literature typically specifies a shield by the property it enforces, though manipulation shields have been evaluated with task performance reported alongside the guarantee~\citep{thumm2022provably}; the case below is a reminder of what an evaluation that omits completion can hide.

\paragraph{Runtime assurance, self-adaptive systems, and models at run time}
The software-engineering literature has already treated most of the pieces this architecture assembles, and the honest position is to say so before claiming anything. The Simplex architecture pairs a high-assurance controller with a high-performance one and switches between them on a monitored condition, so that an unverified component can be deployed behind a verified envelope~\citep{sha2001simplex}; our governance layer is a coarser member of that family, filtering actions rather than switching controllers, and we claim no priority over it. Runtime verification builds monitors from formal specifications and checks executions against them~\citep{leucker2009rv,bartocci2018rv}, and runtime enforcement asks which policies a monitor can correct rather than only observe~\citep{schneider2000enforceable,falcone2011enforcement}; our predicates are a small and informal instance of what that work formalises, and \Cref{sec:exp-govstress} reports what the informal instance misses. Self-adaptive systems research supplies the loop itself: MAPE-K names monitor, analyse, plan, and execute over shared knowledge~\citep{kephart2003autonomic}, and two decades of work have mapped the design space around it~\citep{cheng2009roadmap,salehie2009selfadaptive,weyns2020selfadaptive}. Models at run time is the closest match to what our version registry does, keeping a causally connected model of the running system available for inspection and change while it runs~\citep{blair2009models,bencomo2019models}. Set against this work, the architecture here contributes no new mechanism, and the SE contribution has to be located elsewhere. It is in the unit being governed. A MAPE-K loop reconfigures components whose behaviour is specified; a runtime monitor checks a trace against a property someone wrote down; a models-at-run-time registry holds model elements with declared semantics. A retrained policy offers none of the three, so the analyse step degrades to a statistical comparison on held-out episodes, the enforced property degrades to a geometric envelope over actions, and the registry entry degrades to a hash and a scalar score. Whether those substitutions are usable is an empirical question, and the answer this paper reports is a qualified one: the gate rejects $22.5\%$ of clean candidates, the detector misses three of twelve deployed regressions (all on one seed), and the envelope covers one arm of a two-armed robot.

\paragraph{Self-evolving embodied agent systems}
Two systems released while this work was in review target the same setting and are the closest current comparisons. \citet{liu2026phyagentos} present PhyAgentOS, a self-evolving operating system for embodied agents that decouples cognitive planning from physical execution and layers safety across preflight checks, action bridges, a target-side guard, and heartbeat monitoring. What evolves there is an epistemic memory: verified outcomes are consolidated into reusable knowledge and corrective lessons, explicitly closing the loop \emph{without retraining}. \citet{fu2026capx} present CaP-X, which benchmarks and improves coding agents that write robot control programs, with a training-free harness that matches human-written code on several manipulation tasks and a reinforcement-learning variant that improves success under verifiable rewards. Both establish that the evolving embodied system is now a mainstream object of study, and PhyAgentOS in particular shows that layered enforcement including a target-side guard is an accepted design; we make no novelty claim for execution-boundary enforcement as such. The gap our work addresses is orthogonal and narrower. When what changes is a set of learned module parameters rather than a memory of lessons or a program that can be diffed and re-read, the artifact carries no changelog, cannot be exhaustively unit-tested in the ordinary deterministic sense, and can regress silently. Neither system reports versioning of its evolving artifacts, a promotion gate that admits a candidate only on held-out evidence, or a rollback path once a regression is live. Concurrent work approaches that gap from several sides. ENPIRE and REVOLVE close the loop from real-robot rollouts back into policy training, the first with coding agents that rewrite the training code, and neither reports an admission gate for the retrained policies or a rollback path~\citep{xiao2026enpire,liu2026revolve}. AgenticRobotics does gate promotion on recorded evidence and measures the false-promotion rate of its gate, but the gate chooses checkpoints inside the development loop, and the system reports no way to revert a regression once it is live~\citep{yu2026agenticrobotics}. A survey of post-deployment self-improvement for humanoid robots prescribes a sequence of gates that a candidate update must pass before it commits, inside a human envelope that can roll back to a known-good state, and it notes the lack of a benchmark designed specifically for self-evolving humanoids~\citep{nguyen2026selfevolving}. What we contribute is the part a promotion gate alone does not cover: a rollback path behind the gate, with the gate and the rollback each measured against injected regressions on learned modules. The claim is narrower than it may sound. Each constituent mechanism is old: versioning and staged promotion come from continuous delivery~\citep{savor2016continuous,tarvo2015canary}, an envelope around an untrusted component comes from Simplex~\citep{sha2001simplex} and from shielding~\citep{alshiekh2018safe}, and a run-time model of the deployed configuration comes from models at run time~\citep{blair2009models}. Two further contemporaries frame the same claim space from adjacent angles: Skillware proposes a software ontology and engineering lifecycle for persistent behavioural artifacts~\citep{fan2026skillware}, and VASO targets formally verifiable self-evolving skills for physical agents~\citep{yang2026vaso}, a verification-flavoured relative of the admission problem our gate addresses statistically. What we have not found in any of this work is the set assembled around a learned robot skill module together with a measurement of the price each piece charges in that setting, and that price is the part of this paper we would defend.

\section{Method}
\label{sec:method}

This section presents the architecture in two parts: the formal model of the decision layer and its capability modules (\Cref{sec:agent}--\Cref{sec:composition}), and the three mechanisms built on it, namely the module lifecycle (\Cref{sec:ecm-lifecycle}), runtime governance (\Cref{sec:runtime}), and capability learning (\Cref{sec:learning}).

\noindent\textbf{Scope of the claims.} The ECM abstraction is inherited from the AEROS runtime architecture~\citep{concurrent2026runtime} and restated here only as far as this paper needs it. What we claim is the runtime realisation and the empirical study: versioned module checkpoints, a held-out promotion gate, deployment-level rollback, and measurements of what each costs on a robot simulator. An action-level enforcement boundary is also implemented and described in \Cref{sec:runtime}, but we claim no result for it, since the realisation we evaluate holds task success at zero and \Cref{sec:exp3} rules out a safety claim for it. We do not claim an interface or contract formalism for module compatibility, nor a process model for governing when evolution may occur at the level of an organisation; those are separate concerns and we make no attempt to settle them here.

\label{sec:formalization}

\subsection{Fixed Decision Layer}
\label{sec:agent}

\begin{definition}[Fixed Decision Layer]
The \emph{decision layer} is the tuple
\begin{equation}
  \label{eq:persistent-agent}
  \mathcal{A} = \bigl(\pi_{\mathcal{A}},\; \mathcal{M}_{\mathcal{A}}^{\text{id}},\; \mathcal{M}_{\mathcal{A}}^{\text{ep}}\bigr),
\end{equation}
where $\pi_{\mathcal{A}}$ is a \emph{decision policy} that selects and sequences capability modules, $\mathcal{M}_{\mathcal{A}}^{\text{id}}$ holds its configuration parameters, and $\mathcal{M}_{\mathcal{A}}^{\text{ep}}$ is an \emph{episodic memory} that may grow over time. The architecture holds $(\pi_{\mathcal{A}},\; \mathcal{M}_{\mathcal{A}}^{\text{id}})$ fixed for the lifetime of a deployment; $\mathcal{M}_{\mathcal{A}}^{\text{ep}}$ is free to accumulate execution traces, since it does not alter which module the layer selects.
\end{definition}

\noindent\textbf{The instance we evaluate, and a correction.} \Cref{eq:persistent-agent} describes the decision layer as a policy over states, which is the general form. The configuration evaluated in this paper is a restricted instance of it, and the system's own stored description of that instance is incorrect. By the system's stored record we mean its development documentation, docstrings, and notes, all shipped with the artifact. It calls the configuration a four-phase decomposition with one module per phase of the task. The implementation contains no phase-scoped training of any kind. The \texttt{phase} field of a capability descriptor is read in exactly one place in the codebase, a registry lookup that filters which module is returned, and it never enters training, reward shaping, initial-state selection, or termination. The training entry point wraps the environment in a pass-through whose \texttt{reset} and \texttt{step} forward to the underlying environment unchanged, and it fine-tunes each module with SAC on the full task, from a standard reset, over the full 500-step horizon, trained alone. The four modules are therefore four independently initialised whole-task policies that differ by a label.

\noindent Phases exist only at evaluation. A fixed schedule divides the episode horizon into equal windows: on T6 the horizon is cut into eight windows of 62 steps, and control cycles reach, grasp, lift, place, reach, grasp, lift, place, so each module drives the robot for two windows. Two consequences follow and we state them before any result rests on them. Each module is evaluated on a state distribution it never trained on, since the only states it ever saw were those reachable from a standard reset under its own solo policy. And nothing in this paper tests a phase decomposition, so no result below is evidence for or against one. What we evaluate is a time-sliced multi-policy configuration, and we call it that from here on.

\noindent The module dispatched at a given step is therefore determined by that schedule rather than by a state-conditioned decision: the learned planner is instantiated and held with gradients disabled, but the evaluation path does not consult it. Everything we report about the decision layer concerns a fixed schedule, and our results say nothing about how a genuinely state-conditioned planner would behave under the same lifecycle machinery. The one arm in which a planner is trained and does route actions is AM+rotate (\Cref{sec:exp2}), which is why its drift figures are the informative ones.

We measure adherence to this design with the \emph{policy drift}
$\Delta\pi^{(t)} = \lVert \theta_{\pi}^{(t)} - \theta_{\pi}^{(0)} \rVert_2 / \lVert \theta_{\pi}^{(0)} \rVert_2$, the normalised parameter distance between the deployed decision policy and the one released at $t{=}0$.

\subsection{Embodied Capability Modules}
\label{sec:ecm}

The Embodied Capability Module (ECM) abstraction used throughout this paper is the runtime-level capability unit introduced in the AEROS runtime architecture~\citep{concurrent2026runtime}; we restate its definition here so that the present paper is self-contained, and then focus on the evolution operator that updates ECMs across iterations.

\begin{definition}[Embodied Capability Module]
An \emph{Embodied Capability Module} (ECM) at version~$k$ is a tuple
\begin{equation}
  \label{eq:ecm-tuple}
  c^{(k)} = \bigl(\theta^{(k)},\; \mathcal{I}_c,\; \mathcal{O}_c,\; \phi_c\bigr),
\end{equation}
where $\theta^{(k)}$ denotes learnable parameters (e.g.\ neural network weights, code, or LLM prompts), $\mathcal{I}_c$ and $\mathcal{O}_c$ are the input and output interfaces, and $\phi_c$ is a \emph{capability descriptor} specifying what the module does and when it applies.
\end{definition}

At any time step~$t$ the system maintains a \emph{capability set}:
\begin{equation}
  \label{eq:capability-set}
  \mathcal{C}(t) = \bigl\{c_1^{(k_1)},\; c_2^{(k_2)},\; \ldots,\; c_n^{(k_n)}\bigr\},
\end{equation}
where each $c_i^{(k_i)}$ is the latest version of the $i$-th ECM.

\subsection{State, Decision, and Execution}
\label{sec:execution}

Let $s_t \in \mathcal{S}$ denote the world state at time~$t$. The agent selects an action (or sub-plan) via
\begin{equation}
  \label{eq:action}
  a_t = \pi_{\mathcal{A}}\!\bigl(s_t,\; \mathcal{C}(t),\; \mathcal{M}_{\mathcal{A}}\bigr).
\end{equation}
Each action~$a_t$ is \emph{not} executed directly by~$\mathcal{A}$; instead it is dispatched through the runtime layer~$\mathcal{R}$:
\begin{equation}
  \label{eq:runtime}
  \hat{a}_t = \mathcal{R}\!\bigl(a_t,\; \Phi\bigr),
\end{equation}
where $\Phi$ is a set of safety and policy constraints.

\subsection{Capability Evolution}
\label{sec:evolution}

\begin{definition}[Capability Evolution]
Given experience $\mathcal{D}_t$ at time~$t$, the capability set evolves as
\begin{equation}
  \label{eq:evolution}
  \mathcal{C}(t{+}1) = \textnormal{\textsc{Update}}\!\bigl(\mathcal{C}(t),\; \mathcal{D}_t\bigr).
\end{equation}
The \textnormal{\textsc{Update}} function may employ reinforcement learning, imitation learning, LLM-based code synthesis, or any combination thereof. It operates \emph{only} on capability parameters~$\theta^{(k)}$; the agent~$\mathcal{A}$ is untouched.
\end{definition}

\noindent\textbf{Design invariants.} Two properties follow from the construction above, and we state them as design commitments rather than as results, since each holds by definition once the decision layer is frozen. First, the deployed decision policy is invariant:
\begin{equation}
  \label{eq:identity-invariance}
  \bigl(\pi_{\mathcal{A}}^{(t)},\; \mathcal{M}_{\mathcal{A}}^{\text{id},(t)}\bigr) = \bigl(\pi_{\mathcal{A}}^{(0)},\; \mathcal{M}_{\mathcal{A}}^{\text{id},(0)}\bigr) \quad \text{for all } t,
\end{equation}
so that all change over a deployment is carried by $\mathcal{C}(0) \to \mathcal{C}(1) \to \cdots$. Second, and as a consequence, any change in task performance $J$ must be attributable to a change in the capability set: $J(t_2) \neq J(t_1) \Rightarrow \mathcal{C}(t_2) \neq \mathcal{C}(t_1)$. The engineering value of these invariants is not that they are hard to achieve, but that they give every deployed behaviour a recorded provenance, namely the module version live at the time, which is what makes the versioning and rollback machinery of \Cref{sec:ecm-lifecycle} meaningful. These invariants are not empirical findings, and the only thing left to check is whether an implementation honours what it declares; \Cref{sec:exp-mechanism} reports that audit (\Cref{tab:notation} summarises the notation).

\noindent Provenance is not causal attribution. Knowing which version was live when a behaviour changed does not establish that the version caused the change, since evaluation noise and environment state move as well; establishing causation needs single-module interventions, which the fault-injection study of \Cref{sec:exp-lifecycle} performs and the observational comparisons do not.

\begin{table}[!ht]
  \centering
  \small
  \caption{Summary of notation.}
  \label{tab:notation}
  \resizebox{\linewidth}{!}{%
  \begin{tabular*}{\linewidth}{@{\extracolsep{\fill}}ll@{}}
    \toprule
    Symbol & Meaning \\
    \midrule
    $\mathcal{A}$ & Persistent agent \\
    $\pi_{\mathcal{A}}$ & Decision policy (selects and sequences modules) \\
    $\mathcal{M}_{\mathcal{A}}^{\text{id}}$ & Identity memory (fixed beliefs, goals) \\
    $\mathcal{M}_{\mathcal{A}}^{\text{ep}}$ & Episodic memory (accumulative execution traces) \\
    $c^{(k)}$ & ECM at version $k$ \\
    $\theta^{(k)}$ & Learnable parameters of ECM version $k$ \\
    $\phi_c$ & Capability descriptor (natural-language spec) \\
    $\mathcal{C}(t)$ & Capability set at time $t$ \\
    $\mathcal{R}$ & Runtime governance layer \\
    $\Phi$ & Safety and policy constraint set \\
    $\Delta\pi$ & Normalised $\ell_2$ policy drift \\
    $J(t)$ & Task performance at time $t$ \\
    \bottomrule
  \end{tabular*}%
  }
\end{table}

\subsection{Capability Composition}
\label{sec:composition}

Complex tasks often require composing multiple ECMs. We define a \emph{composition operator}:
\begin{equation}
  \label{eq:composition}
  c_{\text{comp}} = c_i \circ c_j,
\end{equation}
where $\mathcal{O}_{c_i}$ is type-compatible with $\mathcal{I}_{c_j}$. The planner $\pi_{\mathcal{A}}$ selects and orders compositions; the design assigns the runtime an interface-compatibility check, which the released implementation does not contain, and constraint enforcement, which it does (\Cref{sec:runtime}). No result below exercises a compatibility check.

\subsection{ECM Lifecycle}
\label{sec:ecm-lifecycle}

Each Embodied Capability Module follows a four-stage lifecycle: \emph{creation}, \emph{deployment}, \emph{evolution}, and \emph{deprecation}.

\noindent\textbf{What is implemented where.} We separate the design from its current realisation, because the two do not coincide. The runtime we build on carries a per-module keep-best heuristic that compares a retrained module against its own previous reward and reverts on regression. The explicit version registry, the composition-level held-out promotion gate, and deployment-level rollback described below are implemented in a separate experiment driver that orchestrates the evolution loop, rather than inside that runtime. That driver is not part of the current public release and will be added at publication (\Cref{sec:exp-setup}). The measurements in \Cref{sec:exp-lifecycle} are therefore measurements of the mechanism, obtained by running it around the unmodified learner, and not of a shipped runtime feature. Migrating the registry and the gate into the runtime proper is engineering we have not done, and readers should treat the lifecycle contribution as a mechanism with an evaluation rather than as a released component.

\paragraph{Creation}
A new ECM can be instantiated in three ways: (i)~manual specification by a human operator, who provides a skill implementation and a capability descriptor~$\phi_c$; (ii)~LLM-based synthesis, where a language model generates executable code from a natural-language task description; or (iii)~cloning an existing ECM and modifying its parameters. In all cases, the ECM is registered with a unique identifier and an initial version tag~$k{=}0$.

\paragraph{Deployment}
Once created, an ECM is added to the active capability set~$\mathcal{C}(t)$ and becomes available for invocation by the agent's planner~$\pi_{\mathcal{A}}$. The runtime layer validates the ECM's interface compatibility ($\mathcal{I}_c$, $\mathcal{O}_c$) before permitting execution. Each invocation is logged with full execution traces, including inputs, outputs, timing, and any safety interventions.

\paragraph{Evolution}
After deployment, an ECM accumulates execution experience~$\mathcal{D}_t$. When sufficient data is collected (determined by a configurable trigger, e.g.\ a minimum number of episodes or a performance degradation signal), the learning subsystem produces an updated version~$c^{(k+1)}$. The previous version~$c^{(k)}$ is retained in a version registry, enabling rollback if the new version underperforms.

\paragraph{Deprecation}
The registry marks a module superseded when a later version dominates it on all evaluation criteria, or when the task distribution no longer requires the capability; superseded modules stay in the registry for auditability and leave the active set. This is bookkeeping that we implement but evaluate nowhere in this paper, and we claim nothing for it.

\subsection{Runtime Governance}
\label{sec:runtime}

The runtime layer~$\mathcal{R}$ mediates every interaction between the agent and the environment. It serves four functions:

\paragraph{Policy enforcement}
Before any ECM execution, the runtime evaluates the proposed action~$a_t$ against a set of safety and policy constraints~$\Phi$. Constraints are expressed as predicate functions over the current state and proposed action:
\begin{equation}
  \label{eq:policy}
  \mathcal{R}(a_t, \Phi) =
  \begin{cases}
    a_t & \text{if } \forall\, \phi_i \in \Phi:\; \phi_i(s_t, a_t) = \texttt{true}, \\
    a_t' & \text{if a safe modification exists}, \\
    \bot & \text{otherwise (action rejected)}.
  \end{cases}
\end{equation}
Constraints may encode workspace boundaries, force limits, velocity caps, or task-specific invariants. The constraint set~$\Phi$ is defined independently of the ECMs and remains fixed during evolution, so the envelope does not move when a module version does.

\noindent\textbf{What the implementation does, and how it departs from that description.} The realisation we evaluate applies four predicates in sequence: a velocity cap that rescales the translational components $a_{1:3}$, a directional workspace clamp that zeroes outward components when the monitored point is within an $8$\,cm margin of a boundary, a keep-out-zone test, and a magnitude check labelled as a force limit. Only three of the four correspond to the middle line of \Cref{eq:policy}. The keep-out branch is the $\bot$ line and it rejects by substitution: when the monitored point lies inside a configured zone widened by the same $8$\,cm margin, the runtime replaces the proposed action with the zero vector, discarding every dimension of it: on the single-arm tasks all seven dimensions including the gripper command, and on the bimanual task both arms. The zone test reads the monitored point's $x$ and $y$ coordinates only, so a boundary approached purely along $z$ is invisible to it. The fourth predicate is a misnomer we carry over from the configuration file. It thresholds $15\lVert a_{1:3}\rVert$, evaluated after the velocity cap has already rescaled $a_{1:3}$, against a limit expressed in newtons; the quantity is a scaled action magnitude, no contact force is read from the simulator at this point, and with the cap at $1.5$ the check is a second velocity limit under another name. Neither fact is visible in \Cref{eq:policy}, and both drive the results of \Cref{sec:exp3}.

\paragraph{Execution management}
The design calls for the runtime to orchestrate ECM invocation with input marshalling, timeout enforcement, exception handling, and configurable recovery (retry, fallback to an alternative ECM, or safe abort). The released implementation contains none of these paths: an ECM invocation is a direct function call, and the runtime's whole role at execution is the action predicates of \Cref{sec:runtime}. We flag this here rather than let it stand as description, because \Cref{sec:experiments} withdraws claims precisely where description had exceeded implementation.

\paragraph{Trace and logging}
The design calls for a structured trace per execution (plan step, ECM version, input and output state, duration, interventions) feeding the experience dataset~$\mathcal{D}_t$. The released trace records the action-level intervention type and step only, no state or duration, and no trace feeds learning: the experience data the learner actually uses is the SAC replay buffer.

\paragraph{Version gating}
When a new ECM version is produced by the learning subsystem, a gated deployment check decides whether it replaces the incumbent. In the released implementation this lives in two places, neither of them the runtime: each ECM carries a keep-best heuristic that reverts a version whose reward history regresses (\Cref{sec:ecm}), and the held-out evaluation gate is implemented in the lifecycle evaluation driver and measured in \Cref{sec:exp-lifecycle}.

\subsection{Capability Learning}
\label{sec:learning}

The learning subsystem implements the \textnormal{\textsc{Update}} function from \Cref{eq:evolution}. We support three learning modalities, selected per-ECM based on the module's representation:

\paragraph{Reinforcement learning}
For ECMs with differentiable parameters (e.g.\ neural network policies), we apply policy gradient methods using reward signals derived from task outcomes. The reward function is defined over the execution trace:
\begin{equation}
  \label{eq:reward}
  r_t = \alpha \cdot \mathbb{1}[\text{success}] - \beta \cdot t_{\text{exec}} - \gamma \cdot n_{\text{retry}},
\end{equation}
where $\alpha$, $\beta$, $\gamma$ are weighting coefficients, $t_{\text{exec}}$ is execution time, and $n_{\text{retry}}$ is the number of retries. This is the design reward and nothing instantiates it: in the released implementation training optimises the simulator's shaped task reward directly, and $\alpha$, $\beta$, $\gamma$ appear nowhere in the code.

\paragraph{Imitation learning}
\looseness=-1 When expert demonstrations are available (e.g.\ from human teleoperation or a high-performing oracle), ECM parameters are updated via behavioural cloning on the demonstration dataset. This modality is particularly useful for bootstrapping new ECMs or correcting systematic errors. \emph{It never executed in any run reported in this paper.} The behavioural-cloning trainer is implemented, and it will collect scripted demonstrations itself when none are supplied, but the call site that would reach it is guarded on a demonstrations argument that no caller in the codebase ever passes, so the guard is never satisfied and the trainer is never entered. We state this here because the system's stored description of this modality does not match the code.

\paragraph{LLM-based code synthesis}
\looseness=-1 The design intent for ECMs implemented as executable code is that a language model analyses failure traces and proposes code modifications, receiving the current source, a summary of recent failures, and the capability descriptor~$\phi_c$, with version gating admitting only improvements. \emph{This modality is not implemented in the system we evaluate, and the component standing in for it never executed either.} What occupies its place applies heuristic parameter perturbations and never queries a language model. That stand-in also never ran: it fires only once a buffer of failure traces reaches a threshold, the buffer is filled only from a failure-trace argument, and no caller in the codebase passes one, so the buffer is always empty and the threshold is never met. We describe the design because the lifecycle machinery is agnostic to which modality proposes a candidate, and we flag the gap so that nothing below is read as evidence about language-model synthesis.

\paragraph{Hybrid updates}
The design admits evolution steps that combine modalities: a language model proposes a structural change to an ECM's control flow, after which RL fine-tunes the numerical parameters within the new structure. No run reported in this paper combined modalities in this way. In every result below, reinforcement learning is the only modality that executed, whatever the configuration was nominally set to, for the two reasons just given.

\section{Experiments}
\label{sec:experiments}

\subsection{Experimental Setup}
\label{sec:exp-setup}

We evaluate the proposed framework on six tabletop and bimanual manipulation tasks implemented in robosuite~1.4.1 (MuJoCo~2.3.2)~\citep{robosuite2020}. All tasks use a Franka Panda 7-DoF arm (or two arms for T6) controlled in operational-space pose (\texttt{OSC\_POSE}). \Cref{tab:tasks} summarises the task suite; \emph{steps} denotes the number of sequential ECM invocations required for success.

\begin{table}[!t]
  \centering
  \footnotesize
  \setlength{\tabcolsep}{2pt}
  \caption{Manipulation tasks used in evaluation, ordered by composition depth. Success for every task is the underlying robosuite environment's \texttt{\_check\_success}, read at the final step; there are no per-task criteria in the implementation. The environment column is the operative one: in the released code the task definitions forward no per-task construction arguments, so T1, T2 and T5 construct byte-identical default \texttt{PickPlace} environments and share one success criterion.}
  \label{tab:tasks}
  \begin{tabular*}{\linewidth}{@{\extracolsep{\fill}}llcl@{}}
    \toprule
    Task & Description & Steps & Environment \\
    \midrule
    T1: Pick            & Grasp object from table      & 2 & \texttt{PickPlace} \\
    T2: Place           & Place at goal pose           & 3 & \texttt{PickPlace} \\
    T3: Stack           & Stack A on B                 & 4 & \texttt{Stack} \\
    T4: NutAssembly     & Seat nut on peg              & 5 & \texttt{NutAssembly} \\
    T5: PickPlaceMulti  & Place 3 objects into bins    & 6 & \texttt{PickPlace} \\
    T6: TwoArmPegInHole & Bimanual peg insertion       & 8 & \texttt{TwoArmPegInHole} \\
    \bottomrule
  \end{tabular*}
\end{table}

T1--T3 primarily exercise individual ECM capabilities (reach, grasp, lift). T4--T6 are the tasks over which more than one module is scheduled within an episode, so they are where the runtime path that hands control between modules is exercised; T6 additionally requires bimanual coordination. The suite is graded rather than uniformly discriminating, and it is worth saying in advance what each part of it can show. The trained comparisons of \Cref{sec:experiments} are all on T6, and the per-seed logs of each are in the artifact. The other five tasks enter only through the evaluation-only governance stress test of \Cref{sec:exp-govstress}, which needs a task's geometry and not a trained policy.

\noindent\textbf{Training.} We use Soft Actor--Critic (SAC, \texttt{stable-baselines3}~2.8)~\citep{haarnoja2018sac}, 50{,}000 environment steps per ECM per iteration with 12 parallel environments (\texttt{SubprocVecEnv}), batch size~4096, replay buffer $10^6$, $\gamma{=}0.99$, $\tau{=}0.005$. Each task is run for 20 evolution iterations, with 30 evaluation episodes per iteration. These values are attested for the cloud runs by their released configuration.

\noindent\textbf{Stochastic perturbations.} Observations are corrupted with Gaussian noise ($\sigma{=}0.01$\,m) and per-step actuation failures occur with probability $0.05$; both are applied in the environment wrapper and we verified the applying code. Object placements are randomised by the simulator's own placement initialiser. The stored record additionally claims a $\pm5$\,cm object-position perturbation from a wrapper parameter; that parameter is stored and threaded through every constructor and applied nowhere, one more description-versus-code defect, so we withdraw that clause and record the dead parameter (\Cref{sec:provenance}).

\noindent\textbf{Runtime governance} is configured with a 1.5 action-space velocity cap, an axis-aligned workspace box $x,y \in [-0.6, 0.6]$\,m, $z \in [0.6, 1.3]$\,m, a keep-out zone $x,y \in [-0.05, 0.05]$\,m, and a limit of 20 recorded in newtons. When the monitored point is within 8\,cm of a workspace boundary, action components directed outward are zeroed; when it is within 8\,cm of the keep-out zone, the whole action vector is replaced by zero. Two properties of this configuration matter for \Cref{sec:exp3} and we state them here rather than after the numbers. Widening the keep-out zone by the margin gives a $\pm 0.13$\,m box centred on the world origin, which lies inside the region the Panda end-effector has to traverse, so the zone is an obstacle in the middle of the workspace and not a keep-out region at its edge. And the 20\,N figure bounds no contact force: it bounds $15\lVert a_{1:3}\rVert$ evaluated after the velocity cap (\Cref{sec:runtime}), so it can fire at $22.5$ and is a second velocity limit.

\noindent\textbf{Metrics.} We report per-iteration success rate (robosuite \texttt{\_check\_success}, read at the final step of the episode; an episode that reaches the success condition mid-way and loses it by the last step scores zero, which is the property the retention analysis of \Cref{sec:exp-nophase} depends on), average reward, reward variance $\sigma^2$, and agent-policy drift $\Delta\pi$ (normalised $\ell_2$ drift of the agent's decision parameters between the first and last iteration).

\noindent\textbf{Baselines.} We compare five configurations. Their training budgets are \emph{not} equal, and the imbalance is large enough that it governs how the cross-method numbers may be read. Each multi-module arm trains four whole-task modules at 50{,}000 environment steps each per iteration, for 200{,}000 steps per iteration in total, whereas Agent Modification, SkiMo and SPiRL run at their published defaults of 20{,}000 steps per iteration. The multi-module arms therefore receive an order of magnitude more environment interaction per iteration than the monolithic and skill-prior arms. This means that no comparison in \Cref{tab:method-peak-success} between a multi-module arm and a single-module one isolates the architecture: every such contrast is also a $200{,}000$ against $20{,}000$ contrast. \Cref{sec:exp2} reports the one comparison in this paper that is budget-matched, and we restrict the architectural claim to it. Each of the four modules in a multi-module arm is trained on the whole task and is dispatched by a fixed time-slice rotation at evaluation (\Cref{sec:agent}); no module is trained on a sub-problem. The configurations are:
\begin{itemize}[nosep]
    \item \textbf{Ours (Capability Evolution)}: agent~$\mathcal{A}$ fixed; four whole-task ECMs, rotated by the fixed time-slice schedule at evaluation, each evolving every iteration through the hybrid learner. The learner is configured with all three modalities enabled, but only SAC fine-tuning executes, for the reasons given in \S\ref{sec:learning}.
    \item \textbf{Agent Modification (AM)}: a single monolithic PPO policy updated each iteration; no ECMs.
    \item \textbf{Static ECM (S-ECM)}: random-init ECMs that are never updated (lower bound on capability-free performance).
    \item \textbf{SPiRL}~\citep{pertsch2021spirl}: offline skill priors with downstream RL that fine-tunes a skill-conditioned policy.
    \item \textbf{SkiMo}~\citep{shi2023skimo}: jointly learned skill repertoire with a world model for planning in skill space; both skill and model parameters are updated each iteration.
\end{itemize}
Every number in this paper comes from one machine and one preserved environment: a rented cloud workstation (RTX~5090, Ubuntu, Python~3.10) under a virtual environment pinning robosuite~1.4.1, MuJoCo~2.3.2, and stable-baselines3~2.8. The method comparison uses seeds 7, 123 and 2024 for all five methods, and the lifecycle, governance-stress, mechanism, severity, and handoff experiments ran on the same stack; \Cref{sec:provenance} states the scope of what is pinned.

\noindent\textbf{Artifact availability.} The replication artifact accompanies this submission and is published as the repository at \url{https://github.com/s20sc/capability-evolution}. It contains the seven evaluation drivers (lifecycle, governance stress, instrumented rerun, detector severity, the arm keyed \texttt{amphase}, the arm keyed \texttt{nophase}, and the handoff intervention), the per-seed result files behind the numbers these experiments report (\Cref{sec:exp-lifecycle,sec:exp-govstress,sec:exp-nophase,sec:exp-mechanism}), the evaluation checkpoints that the handoff and detector-severity drivers load, with their checksums, the analysis scripts that regenerate \Cref{tab:method-peak-success,tab:retention}, the scripts that render every data figure, a README describing this paper, and dependencies pinned to robosuite~1.4.1, MuJoCo~2.3.2, and stable-baselines3~2.8; the two training-path defects of \Cref{sec:provenance} are corrected in it and the corrections documented in place. Raw console logs, which embed machine-local paths, stay with the authors and are available on request. The artifact also holds the parts of the system that predate this evaluation: the six-task suite and the robosuite wrapper, the ECM data model and the persistent agent, the reinforcement, imitation, and synthesis learner modules, the four published baselines, the runtime governance layer, the driver for the evolution, method-comparison, safety, and ablation experiments, the multi-seed T6 driver, and the aggregate result files those two drivers write. Two provenance facts about that stored code belong here. First, the stored multi-seed driver runs only two of the five methods, so the baseline columns of \Cref{tab:method-peak-success} rest on an extended version of it that ran on the cloud machine, and the per-seed result files for SkiMo, SPiRL and Static~ECM are deposited with the rest. Second, the development-era dependency file declared \texttt{robosuite$\geq$1.4.1} and \texttt{mujoco$\geq$3.0}, which resolve to versions the code cannot run, because the environment wrapper calls \texttt{load\_controller\_config}, an entry point removed after robosuite~1.4.1; the artifact pins the runnable stack instead. The driver also carries the ECM registry and the per-module keep-best revert described in \Cref{sec:ecm-lifecycle}, which are the parts of the lifecycle that live inside the runtime. The lifecycle, governance-stress, mechanism, and handoff experiments are attested to robosuite~1.4.1 with MuJoCo~2.3.2 by the preserved environment they ran in, and \Cref{sec:provenance} states exactly what is pinned.

\FloatBarrier

\subsection{Provenance: Which Code Produced Which Number}
\label{sec:provenance}

\looseness=-1 Reading this system's code against its description has repeatedly produced divergences, so before reporting results we state exactly which code stands behind which number. Where the stored record and the code disagree, we report the code. Two code states are involved: the state that ran on the cloud workstation, which produced every number this paper reports, and the released repository. They differ in one identified place, stated below, and the difference is corrected in the artifact.

\noindent\textbf{The released training path cannot run as pinned.} The released \texttt{\_make\_vec\_env} rebuilds subprocess training environments from a partial configuration that omits the seed, so environment noise, actuation-failure, and object-placement streams are entropy-seeded and the reported ``seeds'' label only network initialisation and SAC sampling; this bounds what a seed controls in every experiment reported here. The same function forwards a \texttt{task\_name} argument that robosuite~1.4.1 environment constructors reject, so under the pinned stack the parallel-training branch raises \texttt{TypeError} on first use. The cloud runs demonstrably trained through that branch, so the executing code differed from the released code at least at this line. Both defects are corrected in the artifact accompanying this submission, with a scope we state exactly: the correction threads distinct per-worker seeds into the wrapper, which controls the wrapper's own streams (observation noise, actuation failures), while the underlying simulator environment still receives no seed, so its internal placement stream remains uncontrolled. A related dead parameter, the wrapper's never-applied object-position perturbation, is recorded in \Cref{sec:exp-setup}.

\noindent\textbf{What is pinned.} The four attested experiment families (\Cref{sec:exp-lifecycle,sec:exp-govstress,sec:exp-nophase,sec:exp-mechanism}) each ran from a driver we release, on the recorded cloud stack, writing per-episode logs we release, and every number in those sections traces to those artifacts. The rule we draw from this section, and the reason it exists, is the same one the rest of the paper keeps arriving at: a description that has not been executed against its artifact is a hypothesis, including the descriptions in experiment notes; the reproducibility literature grades AI results by exactly this gap between what is documented and what is shipped~\citep{gundersen2018reproducibility}.

\FloatBarrier

\subsection{Method Comparison on Composite Tasks (H1)}
\label{sec:exp2}

\looseness=-1 We next compare Ours against the four baselines on T6, the discriminating task, reporting peak success rate over 20 evolution iterations at the three attested seeds (\Cref{tab:method-peak-success}).

\begin{table}[!ht]
  \centering
  \footnotesize
  \setlength{\tabcolsep}{4pt}
  \caption{T6 peak success rate (\%) over 20 iterations for the five methods, per seed and as mean\,$\pm$\,std over the attested seeds $\{7, 123, 2024\}$. Every value comes from a run on the pinned cloud stack, and the per-seed result files are in the artifact (\texttt{analysis/method\_comparison\_n3.py} regenerates this table). We attach no hypothesis test: an exact two-sided test at three seeds per arm cannot return $p$ below $2/20{=}0.1$, and the spread within a single method (Ours spans $23.3$ to $73.3$) is what this table should be read for. A fourth, baseline-only seed (42) exists in the artifact for the three recovered baselines and is excluded here because Ours and Agent Modification were not run on it; including it would raise no method above Ours. The budget-matched comparison this paper's architectural claim rests on is in \Cref{sec:exp-nophase}.}
  \label{tab:method-peak-success}
  \begin{tabular*}{\linewidth}{@{\extracolsep{\fill}}lcccc@{}}
    \toprule
    Method & Seed 7 & Seed 123 & Seed 2024 & Mean$\pm$std \\
    \midrule
    Ours                 & 73.3 & 23.3 & 70.0 & $55.6{\,\pm\,}28.0$ \\
    Agent Modification   & 20.0 & 16.7 &  3.3 & $13.3{\,\pm\,}8.8$ \\
    SkiMo                & 46.7 & 20.0 & 33.3 & $33.3{\,\pm\,}13.3$ \\
    SPiRL                &  0.0 &  0.0 &  0.0 & $0.0{\,\pm\,}0.0$ \\
    Static ECM           &  0.0 &  0.0 &  0.0 & $0.0{\,\pm\,}0.0$ \\
    \bottomrule
  \end{tabular*}
\end{table}

\noindent\textbf{T6: the discriminating task.} On T6, an 8-step bimanual peg-in-hole task, our method reaches a mean peak success of $55.6 \pm 28.0$\% over the three attested seeds (per seed: 73.3, 23.3, 70.0\%), monolithic Agent Modification reaches $13.3 \pm 8.8$\% (20.0, 16.7, 3.3\%), SkiMo, which jointly evolves its skills and planner, reaches $33.3 \pm 13.3$\% (46.7, 20.0, 33.3\%), and SPiRL and Static~ECM never solve the task on any seed. We attach no test to these gaps, for two reasons stated in the caption: an exact two-sided test at three seeds per arm has a floor of $p{=}2/20{=}0.1$, and the within-method spread is the more instructive number anyway, since our own method's seeds span 23.3 to 73.3\% at identical configuration, a factor of three from the random stream alone. That spread is this table's finding: single-run peak success cannot rank these methods at seed counts a study like this can afford. The row also invites the reading that splitting a long-horizon task across modules is what beats monolithic updating and frozen skill priors. That reading fails twice over: the contrast is also a budget contrast, which \Cref{sec:exp-nophase} settles against it, and the configuration on the winning side was never a split of the task at all (\Cref{sec:agent}).

\noindent\textbf{Per-seed variance on T6.} Two of the three Ours seeds yield peak success at 70.0 and 73.3\%; seed~123 stays below 14\% for nineteen iterations and reaches only 23.3\% on the twentieth, consistent with slow policy bootstrap and suggesting that the 20-iteration budget is near the lower edge of what T6 requires. We report the full three-seed mean.

\smallskip
\noindent\textbf{Does the frozen agent cost anything?} The comparison so far pairs a four-policy, frozen-agent method (Ours) against a single-policy, agent-mutating one (Agent Modification), which leaves the frozen agent confounded with everything else that differs. To isolate it we run a matched baseline, \emph{AM+rotate}, given the same four whole-task ECMs and the same SAC learner as Ours, with the imitation and synthesis paths equally inert in both arms (\S\ref{sec:learning}), but with its dispatcher free to co-train and to route actions at evaluation, so its agent drifts (mean $\Delta\pi{=}2.36$ across the attested seeds by the drift metric of \Cref{sec:exp-setup}, versus $\Delta\pi{=}0$ for Ours). AM+rotate and Ours differ in exactly one factor, a mutable versus a frozen dispatcher, given identical structure and learning signal; our run logs key this arm as \texttt{amphase}. Over the same three seeds AM+rotate reaches a mean peak success of $55.6 \pm 15.8$\% (per seed: 50.0, 73.3, 43.3), far above monolithic Agent Modification ($13.3 \pm 8.8$\%) and the strongest prior baseline SkiMo ($33.3 \pm 13.3$\%), and its mean coincides with Ours ($55.6 \pm 28.0$\%) to one decimal. We attach no test here either, and none is needed for the direction of the conclusion: the two point estimates are equal, so freezing the agent adds no separable performance advantage and costs none we can resolve; its value is the by-construction identity-invariance ($\Delta\pi{=}0$; \Cref{sec:identity}), so the stability, auditability, and rollback benefits of a frozen decision layer come at no measurable price at this sample size. We do not read this as proven equivalence: three seeds cannot resolve a modest true difference in either direction, and the two arms' spreads differ (\Cref{tab:matched}).

\noindent The other contrast this pairing offers, AM+rotate against monolithic Agent Modification, is arithmetically the largest in the study ($+42.2$ points at the attested seeds). It is neither a decomposition effect nor an architectural one. The two arms differ in the number of whole-task policies being trained and hence in interaction budget, $200{,}000$ steps per iteration against the monolithic arm's published default, and \Cref{sec:exp-nophase} spends the larger budget on one policy and reverses the ordering. We therefore report the contrast and claim nothing from it.

\begin{table}[!ht]
  \centering
  \footnotesize
  \setlength{\tabcolsep}{3pt}
  \caption{The T6 comparison across the four configurations (peak success \% over the attested seeds $\{7,123,2024\}$). The two right-hand columns spend $200{,}000$ interaction steps per iteration against the monolithic column's published default, so the table confounds structure with compute, and the budget-matched control of \Cref{sec:exp-nophase} reverses the ordering. The structure itself is four whole-task policies taking turns, not a split of the task (\Cref{sec:agent}). The frozen and unfrozen variants have equal mean peak success; they differ in drift, which is the factor under test.}
  \label{tab:matched}
  \begin{tabular*}{\linewidth}{@{\extracolsep{\fill}}lcccc@{}}
    \toprule
     & AM & SkiMo & AM+rotate & Ours \\
    \midrule
    Four rotating whole-task policies & --- & skill prior & \checkmark & \checkmark \\
    Frozen agent ($\Delta\pi{=}0$) & --- & --- & --- & \checkmark \\
    \midrule
    T6 peak (\%) & $13.3{\,\pm\,}8.8$ & $33.3{\,\pm\,}13.3$ & $55.6{\,\pm\,}15.8$ & $55.6{\,\pm\,}28.0$ \\
    Agent drift $\Delta\pi$ & 1.55 & 1.34 & 2.36 & 0 \\
    \bottomrule
  \end{tabular*}
\end{table}

\subsection{Conformance Check: Is the Freeze Implemented?}
\label{sec:identity}

\looseness=-1 This subsection reports a conformance check, not a result. That a frozen decision layer exhibits no parameter drift is true by construction, and we do not present it as a finding; a by-construction property billed as evidence would be circular. What is worth checking is whether the implementation honours the design, since a freeze that silently leaks gradients would invalidate every attribution the architecture claims to support. We therefore measure $\Delta\pi$, the normalised $\ell_2$ drift of the decision-policy parameters between the first and last iteration, and read it only as an implementation check. The informative comparison is with the configurations that permit mutation, which quantifies how much a decision layer moves when nothing holds it in place.

\begin{figure}[!ht]
  \centering
  \includegraphics[width=0.95\linewidth]{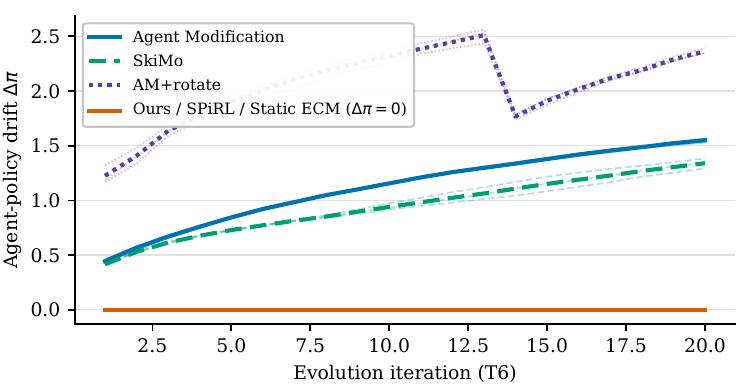}
  \caption{Agent-policy drift $\Delta\pi$ over the 20 T6 evolution iterations, thin lines per attested seed $\{7,123,2024\}$ and thick lines seed means, rendered by \texttt{figures/render\_fig\_identity\_drift.py} from the released per-seed files. Ours, SPiRL, and Static~ECM hold $\Delta\pi{=}0$ by construction; Agent Modification ends at a mean of $1.55$, SkiMo at $1.34$, and the mutable-dispatcher arm AM+rotate at $2.36$.}
  \label{fig:identity-drift}
\end{figure}

Ours, SPiRL, and Static~ECM hold $\Delta\pi{=}0$ on all six tasks by architectural construction rather than as an empirical finding, since each of these methods never writes to the decision policy. The one thing an experiment can add here is an audit that the implementation does not violate what it declares, and \Cref{sec:exp-mechanism} reports that audit on T6: the decision layer's output on a fixed 100-state probe set is byte-identical across all 20 iterations of both instrumented seeds (that experiment runs two), confirming no accidental write to parameters that are held with gradients disabled. Agent Modification and SkiMo exhibit substantial drift, ending the T6 runs at means of $\Delta\pi{=}1.55$ and~$\Delta\pi{=}1.34$ respectively over the attested seeds (\Cref{fig:identity-drift}; qualitative comparison in \Cref{tab:qualitative}). This measures the identity dimension of each method, not the performance driver. The matched AM+rotate baseline (\S\ref{sec:exp2}) shows that agent drift by itself does not decide the T6 outcome, since a drifting dispatcher over the same four whole-task policies reaches the same $55.6\%$ mean peak as the frozen one. What the freeze buys is provenance rather than performance, and the budget-matched control of \Cref{sec:exp-nophase} shows that the structure the freeze was paired with is the wrong one for this task. The conformance result is unaffected by that: a decision layer held fixed is what makes the version live at a given moment the only thing that can have changed, whichever module structure it dispatches.

\begin{table}[!ht]
  \centering
  \footnotesize
  \setlength{\tabcolsep}{2pt}
  \caption{Skill-management properties across frameworks. The composition row refers to scheduling more than one module within a single episode, which T4--T6 exercise through the fixed time-slice rotation of \Cref{sec:agent}; the type-compatibility checking of \Cref{sec:composition} is implemented but not independently evaluated. The runtime-safety row records that an enforcement layer is present, not that ours is usable: \Cref{sec:exp3} shows it holds task success at zero. Voyager rewrites a skill in place when it is regenerated and keeps the earlier code as numbered files, with no mechanism for choosing or restoring a version, hence partial versioning and no rollback.}
  \label{tab:qualitative}
  \begin{tabular*}{\linewidth}{@{\extracolsep{\fill}}lccccc@{}}
    \toprule
    Property & SPiRL & SkiMo & Voyager & AM & Ours \\
    \midrule
    Skill versioning     & \texttimes & \texttimes & partial & --- & \checkmark \\
    Continuous evolution & \texttimes & \checkmark & \checkmark & \checkmark & \checkmark \\
    Rollback support     & \texttimes & \texttimes & \texttimes & \texttimes & \checkmark \\
    Agent identity fixed & \checkmark & \texttimes & \texttimes & \texttimes & \checkmark \\
    Runtime safety layer & \texttimes & \texttimes & heuristic & \texttimes & \checkmark \\
    Composition tested   & \texttimes & \checkmark & \texttimes & \texttimes & \checkmark \\
    \bottomrule
  \end{tabular*}
\end{table}

\subsection{The Shield Scored Perfectly and Stopped the Robot (H2)}
\label{sec:exp3}

\looseness=-1 The runtime layer of \Cref{sec:runtime} screens every proposed action. Read against its own code and then driven with real end-effector state (\Cref{sec:exp-govstress}), it fails as a safety mechanism in a way that a violation-count acceptance criterion cannot see, and this subsection reports that failure as a negative result about evaluation design, which we think is the more useful of the two things we could have written here.

\noindent\textbf{Two defects, and they are different in kind.} The first is a configuration error. The keep-out zone is set to $x,y \in [-0.05, 0.05]$\,m, the enforcement code widens every zone by its 8\,cm margin, and the result is a $\pm 0.13$\,m box centred on the world origin, which lies within the volume a Panda end-effector has to move through. Nothing about the mechanism is wrong here; the zone was placed on top of the workspace instead of outside it. The keep-out branch is also the one predicate that does not modify an action. It substitutes the zero vector for the whole action, all 12 dimensions on T6, both arms (this task's action space carries no gripper dimensions; \Cref{sec:runtime}), so an intervention on this branch is not a correction but a refusal to act.

The second is a wiring error, and it is what makes any violation-count reading of this layer uninterpretable rather than merely unflattering. The evaluation loop obtains the end-effector position with \texttt{info.get("ee\_pos", np.zeros(3))} after each environment step, and the environment wrapper never writes \texttt{ee\_pos} into \texttt{info}. The wrapper does read the end-effector position, as \texttt{robot0\_eef\_pos} from the observation dictionary, and uses it for its own violation check, but it keeps it there. Under the released code the default is therefore taken at every step, the shield evaluates its predicates at the origin, a point the arm never occupies, and since the origin sits inside the widened keep-out box, the keep-out branch fires unconditionally and blocks every action. Driving the same enforcement code with verified real positions (\Cref{sec:exp-govstress}) removes this defect and reproduces the paralysis anyway. The same zero vector is written as a literal in the training-time action wrapper, so that path would have had the same defect had it ever been given a runtime.

\noindent\textbf{Why the instrument could not see it.} The driver for this experiment records violations, blocked actions, and overhead. It records no success rate. A shield that returns the zero vector for every action earns a perfect violation count and a small overhead, and there was no channel on which that would have looked like a problem. The acceptance criterion contained the property to be enforced and nothing about the system that was supposed to keep working under it.

\noindent\textbf{The cost, measured.} \Cref{sec:exp-govstress} drives the same enforcement code from real end-effector positions, so the predicates there see the state the robot is in. The outcome does not improve. The keep-out branch still fires on 75 to 100\% of steps depending on the task, because the zone placement alone is sufficient, and task success under governance is $0.0\%$ on all six tasks. On T6, the only task in that experiment with a trained checkpoint and a nonzero baseline, success falls from $2.0\%$ with governance off to $0.0\%$ with it on.

\noindent\textbf{H2 fails, and we state what replaces it.} We do not claim that the runtime layer enforces a safety envelope at negligible cost. What we report instead is a measurement-validity failure with a shape that generalises past this system, though we should be precise about what generalises: this is one self-diagnosis rather than a survey of how runtime shields are evaluated, so what transfers is the analytic point about acceptance criteria and not an empirical claim about how often the trap is sprung. A safety mechanism whose acceptance criterion omits task completion will score perfectly by disabling the thing it protects, and it will do so silently, since every number the criterion contains moves in the direction the designer wants. The two defects underneath are worth separating because they call for different fixes and different suspicions. A keep-out zone overlapping the workspace is a configuration error, catchable by asserting that the configured envelope admits a known-good trajectory. Predicates evaluated at a point the robot never occupies is a wiring error, catchable by asserting that the state handed to the shield is the state the environment reports. Neither assertion existed here, and no metric we collected would have raised either. The general form of the lesson is the one we draw again in \Cref{sec:discussion}: a mechanism claim needs evidence that the mechanism fired and evidence that the system still works, and a violation-count criterion collects only the first.

\subsection{Lifecycle Mechanisms: Gated Promotion and Rollback (H3)}
\label{sec:exp-lifecycle}

The mechanisms of \Cref{sec:ecm-lifecycle} are only worth their complexity if they change what reaches deployment. We test them by injecting regressed module versions into the evolution loop on T6 and comparing three deployment policies over the seeds $\{42,7,123,2024\}$ and 13 iterations each.

\noindent\textbf{Setup.} At iterations $\{5,8,11\}$ the candidate produced by that iteration's training is replaced by a regressed version, obtained by adding Gaussian noise to every module's parameters with a per-tensor scale of $0.5$ standard deviations. Three arms differ only in what the deployment layer does with a candidate. \emph{Gate-off} promotes every candidate unconditionally. \emph{Gate-on} evaluates candidate and incumbent on 30 fresh held-out episodes each and promotes the candidate only if its mean episode reward stays within 10\% of the incumbent's; the gate metric is dense reward throughout, not binary success. \emph{Rollback} promotes unconditionally, then restores the previous version whenever deployed performance falls more than 15\% below the median of the last three iterations. All three arms retain the base learner's own per-module keep-best heuristic, so the arms isolate the deployment-level machinery rather than replacing the learner, which is implemented in the experiment driver rather than in the runtime (\Cref{sec:ecm-lifecycle}). One caveat: after the driver restores a snapshot, a module's internal keep-best history is not rebuilt, so its reference reward can point at a replaced candidate; the driver accepts this drift and we record it. The injection scale was calibrated to make a regressed version cost well above the gate's 10\% threshold; the twelve measured held-out costs run from $16.0$ to $36.4\%$, four of them below the $20\%$ the calibration targeted and all comfortably above the threshold; a rejection rate of 100\% is therefore close to what the calibration guarantees, and the informative quantities in this experiment are the clean-candidate rejection rate, the size of the dip when nothing intercepts the regression, and whether rollback fires without cause. We call a candidate \emph{clean} when no noise was injected into it; a clean candidate is not thereby known to be an improvement, since natural training can itself regress, so this rate is a rejection rate on unlabelled candidates and not a false-positive rate against ground truth. We report the dense episode reward here because T6 binary success is a late, heavy-tailed event (\Cref{tab:method-peak-success}); an injected regression that is plainly visible in reward can be invisible in a success rate that is zero both before and after.

\noindent\textbf{This experiment is not contaminated by the shield.} \Cref{sec:exp3} refutes H2, and it is worth saying which results that reaches. The training path constructs its vectorised environments through a factory that wraps each one in an action wrapper without passing a runtime, so the enforcement code has no handle to call and never executes inside a training subprocess. The governance layer was therefore inert throughout the runs reported in this subsection and in \Cref{sec:exp2,sec:exp-nophase,sec:exp-mechanism}, which are consequently unaffected by the defects described above. The irony is worth recording: a layer we described as mediating every interaction between the agent and the environment did not run during any of the training that produced the modules it was meant to govern.

\begin{table}[!ht]
  \centering
  \footnotesize
  \caption{Lifecycle mechanisms on T6 under injected regressions, over four seeds and three injections per run. Dip depth is the drop at the injected iteration itself, measured against the mean of the three preceding iterations; the rollback trigger separately uses the median of the last three (\Cref{sec:exp-lifecycle}). The released aggregate file also records a windowed variant that takes the deepest drop anywhere in the interval following an injection. The two conventions agree on all twelve gate-off episodes and on eleven of twelve rollback episodes, but diverge on seven of twelve gate-on episodes, where the windowed statistic reads $0.7\pm3.9$\% against the at-injection $-1.0\pm4.3$\% printed here: with nothing to intercept, the deepest drop in a window is noise, whereas the injected iteration itself sits marginally above its own baseline. Dip and recovery statistics pool the twelve injection episodes, which cluster three-per-seed and are not independent; per-seed mean dips are $0.9, -0.5, -1.2, -3.4$\% under the gate (seed-level $-1.0\pm1.8$), $16.8, 18.9, 7.7, 14.4$\% without it (seed-level $14.5\pm4.9$) and $19.3, 17.2, 21.9, 12.2$\% under rollback (seed-level $17.6\pm4.1$), so the seed-level and pooled means agree on all three arms; the dispersions differ because one aggregates twelve episodes and the other four seeds. All dispersions in this table and caption are sample standard deviations. Clean-candidate rejection is computed per seed. Gate-on additionally spends two held-out evaluations of 30 episodes per iteration.}
  \label{tab:lifecycle}
  \begin{tabular*}{\linewidth}{@{\extracolsep{\fill}}lccccc@{}}
    \toprule
    Arm & Regressions & False & Dip & Recovery & Rollbacks \\
        & rejected    & rejections & depth & (iters) & (false) \\
    \midrule
    Gate-on   & $100\pm0$\% & $22.5\pm26.3$\% & $-1.0\pm4.3$\% & $0.0\pm0.0$ & n/a \\
    Gate-off  & n/a & n/a & $14.5\pm8.1$\% & $0.9\pm0.3$\textsuperscript{\dag} & n/a \\
    Rollback  & n/a & n/a & $17.6\pm7.2$\% & $0.9\pm0.3$ & $2.2\pm1.5$ (0) \\
    \bottomrule
  \end{tabular*}
  \vspace{2pt}
  {\footnotesize \textsuperscript{\dag}One of the twelve injection episodes never returned to baseline within its window and is censored; the mean is over the eleven that did.}
\end{table}

\begin{figure}[!t]
  \centering
  \includegraphics[width=\linewidth]{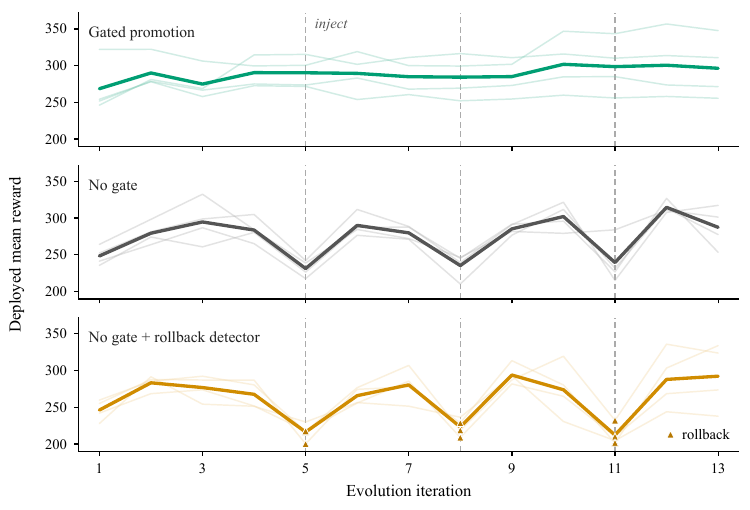}
  \caption{Deployed mean reward under injected regressions on T6, per deployment policy. Thin lines are the four seeds, thick lines their mean; dashed verticals mark the injection iterations $\{5,8,11\}$. Top: gated promotion intercepts every regressed candidate and the deployed curve is flat through the injections. Middle: without the gate the same regressions reach deployment and cost $14.5\pm8.1$\% of reward. Bottom: the dip detector reverses deployed regressions where it fires (triangles); seed 2024 fired zero times and carries no markers. Rendered by \texttt{figures/render\_fig\_lifecycle.py} in the artifact from the released per-seed result files.}
  \label{fig:lifecycle}
\end{figure}

\noindent\textbf{Results.} \Cref{tab:lifecycle} and \Cref{fig:lifecycle} separate the three policies. Gated promotion rejected all twelve injected regressions, as the calibration anticipated, and the deployed curve shows no systematic drop at the injection points (mean dip $-1.0\pm4.3$\%, that is, marginally above the preceding baseline on average, with five of the twelve episodes showing a small positive dip). The result to read here is not the rejection count, which the calibration nearly guarantees; it is the pair of measured quantities, the absence of a deployment drop and the clean-candidate rejection rate below. Without the gate the same regressions reach deployment and cost $14.5\pm8.1$\% of reward, recovered after roughly one further iteration of training in eleven of the twelve injection episodes; the twelfth never returned to baseline within its window and is censored in that mean. Rollback detects and reverses the deployed regressions with no false triggers on clean iterations, but the average of $2.2$ of $3$ injections caught per run conceals its distribution: the per-seed counts are $3$, $3$, $3$ and $0$, so three seeds caught every injection and one caught none, its dips staying within the 15\% band under the detector's own median-of-last-three baseline. The failure mode is not a uniform tax but an occasional run on which the detector is blind throughout, and because the misses cluster in one run, the twelve injections are four run-level observations rather than twelve independent trials. The comparison is therefore not gate versus rollback but gate before rollback, since the held-out check inspects a candidate that has not yet been exposed to the deployment, whereas the dip detector can only react to one that has.

\noindent\textbf{Detector operating characteristic.} To characterize the detector beyond the three calibrated injections, we swept injection severity on each seed's final incumbent modules, evaluation-only: $\sigma \in \{0, 0.05, 0.1, 0.15, 0.2, 0.35, 0.5\}$, three fresh perturbation replicates per cell, 30 deployment episodes each, scored by the same rule against the same baseline statistic the in-loop detector keeps (\Cref{tab:severity}). This is a characterization on frozen incumbents, not a re-run of the in-loop experiment. Three properties emerge. Specificity is clean: none of the twelve unperturbed re-evaluations triggers. Sensitivity is a threshold contract, not a detection rate: no seed triggers at any $\sigma \le 0.2$, because those regressions cost less than the $15\%$ the rule is set to tolerate, so sub-threshold regressions pass silently by design and any deployment relying on this detector accepts that class of loss. And the in-loop miss cluster has a mechanism: at $\sigma{=}0.5$ two seeds sit near the trigger line rather than clearly above it, seed 42 at $17.7 \pm 3.3$\% and seed 2024 at $18.3 \pm 4.8$\% (one of whose three replicates falls below it), against $22.7$--$27.5$\% for seeds 123 and 7. Which near-threshold seed misses in a given batch is then a matter of where its dips land: in the in-loop run, seed 42's dips cleared the line and seed 2024's all fell inside the tolerance band, at $9.5$, $6.0$ and $13.4\%$ measured as the detector measures them, against the median of its last three deployed rewards. Under the mean-of-preceding-three convention that \Cref{tab:lifecycle} aggregates, the third of those injections reads $19.2\%$; the two figures differ in baseline convention, not in whether a trigger was missed, and we state both so neither can be mistaken for the other. The detector did not fail randomly on one run; a fixed relative threshold is systematically fragile on modules whose degradation sits near it, and the per-seed severity response, not the aggregate catch count, is the quantity that predicts where the misses will be.

\begin{table}[!ht]
  \centering
  \footnotesize
  \caption{Detector severity sweep on frozen final incumbents. Each cell: three perturbation replicates, 30 episodes each; drop is relative to that run's own baseline (median of its last three clean deployed rewards); trigger applies the deployed rule (drop $>15\%$). The range column spans the four per-seed means, not individual replicates, whose spread is wider (the deepest single replicate at $\sigma{=}0.2$ reaches $13.6\%$, still below the trigger). The $\sigma{=}0$ row is clean re-evaluation and estimates the false-trigger rate. Produced by \texttt{severity\_runner.py}; the driver, the per-cell outputs, and the module weights it evaluates are all in the artifact.}
  \label{tab:severity}
  \begin{tabular*}{\linewidth}{@{\extracolsep{\fill}}lcc@{}}
    \toprule
    $\sigma$ & Mean drop \% across seeds (range) & Triggers of 3, seeds $7/42/123/2024$ \\
    \midrule
    $0$    & $-10.3$ to $-0.5$ & 0\,/\,0\,/\,0\,/\,0 \\
    $0.05$ & $-7.7$ to $0.9$   & 0\,/\,0\,/\,0\,/\,0 \\
    $0.1$  & $-3.0$ to $4.2$   & 0\,/\,0\,/\,0\,/\,0 \\
    $0.15$ & $1.7$ to $7.1$    & 0\,/\,0\,/\,0\,/\,0 \\
    $0.2$  & $1.1$ to $12.7$   & 0\,/\,0\,/\,0\,/\,0 \\
    $0.35$ & $10.9$ to $21.7$  & 3\,/\,0\,/\,2\,/\,0 \\
    $0.5$  & $17.7$ to $27.5$  & 3\,/\,3\,/\,3\,/\,2 \\
    \bottomrule
  \end{tabular*}
\end{table}

\noindent\textbf{Cost.} Gating is not free. It rejected $22.5\pm26.3$\% of clean candidates, roughly one non-injected candidate in five, and it consumes two extra held-out evaluations per iteration. Whether those rejected candidates were genuinely better than their incumbents is not something this design labels, so we do not call the rate a false-positive rate or the rejections discarded improvements. The wide standard deviation reflects that the rate is seed-dependent ($0$, $0$, $40$ and $50$\% across seeds). A deployment that values throughput over stability would loosen the threshold; we did not tune it per seed.

\subsection{Governance under Adversarial Probing}
\label{sec:exp-govstress}

The paralysis of \Cref{sec:exp3} follows from reading the layer's code, and a code reading cannot say what the layer does when a robot actually approaches its envelope. This subsection measures that. It challenges the envelope deliberately with the same enforcement code, driven from the end-effector position the simulator actually reports, and it records whether the task was completed.

\noindent\textbf{Setup.} We run 50 episodes per task with governance enabled and disabled, injecting scripted unsafe actions in bursts. The policy driving those episodes differs by task and this matters for how the results should be read: T3 and T6 replay trained module checkpoints found on the evaluation machine (the T3 one recorded as \texttt{multiseed\_T3\_Stack\_seed\_123\_ours}, from a multi-seed run outside the released drivers), while T1, T2, T4 and T5 had none there and are driven by randomly initialised modules. The two replayed checkpoints were not preserved: the per-episode logs of all six rows are in the artifact, but those two rows cannot be replayed with their original weights. The probes and the enforcement path are identical either way, so the four random-policy tasks still exercise the envelope, but they do not tell us how the envelope behaves under a competent policy, and we mark them as such throughout. Three families of probe correspond to three of the four predicates the layer implements: a velocity command beyond the cap, a translation directed at the nearest workspace bound, and a downward force spike. Magnitudes were calibrated on T1 until the disabled condition produced violations in every episode. Because the simulator's own violation check covers workspace bounds, keep-out zones, and contact force but not commanded velocity, blocking a velocity probe is true by construction; we therefore report the violation reduction computed without the velocity family alongside the pooled figure. End-effector extraction succeeded on all $25{,}050$ attempts per task, a count each released result file records in its metadata (\texttt{ee\_extraction\_failures} $=0$ of \texttt{ee\_extraction\_total} $=25{,}050$), so the predicates here saw real positions.

\begin{table}[!ht]
  \centering
  \footnotesize
  \setlength{\tabcolsep}{2pt}
  \caption{Environment-verified violations, task success, and keep-out interventions under adversarial probing, 50 episodes per task per condition. Reduction excluding the velocity family is given in parentheses. The policy column records whether the episodes were driven by trained module checkpoints or by randomly initialised modules; only T3 and T6 had checkpoints available, and only T6 has a nonzero success rate to lose. ``Discarded'' is the fraction of the $25{,}000$ actions per task on which the keep-out branch replaced the whole action vector with zero.}
  \label{tab:govstress}
  \resizebox{\linewidth}{!}{%
  \begin{tabular}{@{}ll r r l c c@{}}
    \toprule
    Task & Policy & Viol.\ (off) & Viol.\ (on) & Reduction & Succ.\ off/on & Discarded \\
    \midrule
    T1: Pick             & random  & 7806  & 0    & 100.0\% (100.0\%) & 0.0 / 0.0\% & 100.0\% \\
    T2: Place            & random  & 9165  & 70   & 99.2\% (99.3\%)   & 0.0 / 0.0\% & 97.9\% \\
    T3: Stack            & trained & 10072 & 0    & 100.0\% (100.0\%) & 0.0 / 0.0\% & 99.8\% \\
    T4: NutAssembly      & random  & 8469  & 152  & 98.2\% (98.2\%)   & 0.0 / 0.0\% & 98.0\% \\
    T5: PickPlaceMulti   & random  & 8847  & 0    & 100.0\% (100.0\%) & 0.0 / 0.0\% & 100.0\% \\
    T6: TwoArmPegInHole  & trained & 6099  & 3973 & 34.9\% (34.0\%)   & 2.0 / 0.0\% & 75.1\% \\
    \bottomrule
  \end{tabular}%
  }
\end{table}

\begin{figure}[!t]
  \centering
  \includegraphics[width=\linewidth]{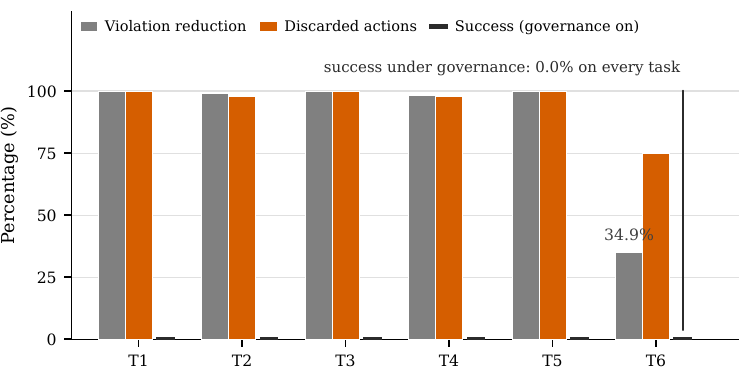}
  \caption{What the shield buys and what it costs, per task. Grey bars: environment-verified violation reduction. Orange bars: fraction of actions whose entire vector the keep-out branch discards. Flat markers: task success under governance, $0.0\%$ on every task. The reduction is bought by the discard, and T6's $34.9\%$ marks where predicates that watch one arm of two stop covering the probes. Rendered by \texttt{figures/render\_fig\_govstress.py} from the released result files.}
  \label{fig:govstress}
\end{figure}

\noindent\textbf{The violation columns, read alone.} \Cref{tab:govstress} tabulates the six tasks and \Cref{fig:govstress} plots the per-task reduction. On the five single-arm tasks the envelope holds under deliberate attack, removing $98.2$ to $100$\% of the violations the same probes cause when governance is disabled, at a per-action cost of $0.025$ to $0.029$\,ms. Excluding the velocity family changes these figures by at most $0.1$ percentage point, so the result does not rest on the one family the simulator cannot independently verify. Read on their own these two columns say the shield works, and they are the columns a violation-count evaluation records.

\noindent\textbf{The success column, which changes what those numbers mean.} No task is completed under governance. Success is $0.0\%$ with the layer enabled on all six tasks, and on T6, the one task with a trained checkpoint and a nonzero baseline, it falls from $2.0\%$ to $0.0\%$. Mean episode reward falls with it, most visibly on the dense-reward T6, from $318.1$ to $299.1$. The last column gives the mechanism. The keep-out branch, which replaces the entire action vector with zero, fires on $75.1$ to $100$\% of the $25{,}000$ actions in each task, and it dominates every other intervention type everywhere: on T1 and T5 it is the only intervention that occurs at all, on T2 and T4 it outnumbers the rest by more than a hundred and fifty to one, and even on T6, where it is least active, it accounts for $18{,}768$ interventions against $1{,}280$ force-limit and $756$ workspace-bound ones. The stored description of this layer, that most interventions are small boundary-margin corrections, is the opposite of what the layer does. Most interventions are refusals to act, and the violation reduction is a consequence of the robot barely moving rather than of the probes being deflected. The zone placement alone is sufficient to produce this: these episodes fed the predicates real end-effector positions, and the outcome matches the run in which they were fed the origin.

\noindent\textbf{Where the envelope stops observing.} On T6 the layer removes only $34.9$\% of violations, and $3973$ crossings survive. Two smaller residuals appear on T2 (70) and T4 (152). The T6 shortfall has a structural cause: the governance predicates are evaluated against the end-effector state of the first robot, so probes that drive the second arm out of bounds are not seen, and contact forces arising between the two arms are outside the monitored envelope entirely. An action-level shield of this design is bounded by what it observes, and a bimanual platform needs a predicate per effector. We keep this observation because it is a real property of the design and because it will matter to anyone building the same kind of layer. It carries no weight in the present system's favour, since a shield that also removes all task performance has no coverage figure worth defending, and it disposes of the invariant-envelope guarantee stated alongside it in the stored documentation: with $3973$ residual crossings on T6 and nonzero residuals on two single-arm tasks, the envelope is not one the layer holds invariant.


\subsection{A Budget-Matched Control: Does the Rotation Pay for Itself?}
\label{sec:exp-nophase}

Every comparison reported so far shares a defect. The four-policy configurations train four whole-task modules at $50{,}000$ environment steps each per iteration, a total of $200{,}000$, while the monolithic baselines train one policy at $20{,}000$ (\texttt{baselines/agent\_modification.py} and \texttt{baselines/skimo\_baseline.py} default to this value and no call site overrides it). Compute differs by an order of magnitude in favour of the design we were advocating, so none of those comparisons can tell whether the structure helps or whether extra interaction does. This subsection reports the control that answers it, and then reports something the control led us to check, which is what the structure under test had been all along.

\noindent\textbf{What is under test, and how the stored description gets it wrong.} The system's stored description calls the four-module configuration a phase decomposition and describes each module as covering one phase of the task. That description does not match the code, and the mismatch is not a matter of emphasis. No training in this system is scoped to a phase: the descriptor field naming a phase is read once in the runtime system, in a registry lookup, and never reaches the modules' training environment, reward, initial state, or termination (the evaluation drivers read the field three further times, for schedule labelling and checkpoint keying, which is enumerable in the artifact); the training wrapper forwards \texttt{reset} and \texttt{step} to the environment untouched; and each module is fine-tuned on the whole task, from a standard reset, over the whole 500-step horizon, alone (\Cref{sec:agent}). The four modules are four independently initialised whole-task policies with different labels. Phases enter only at evaluation, where the horizon is cut into eight windows of 62 steps and control cycles through the four modules twice. What this control tests, therefore, is rotating control among independently trained whole-task policies on a fixed schedule, against giving the whole budget to one such policy.

\noindent\textbf{Setup.} We add an arm, \emph{AM+single} (keyed \texttt{nophase} in our run logs), that trains a single whole-task module at $200{,}000$ steps per iteration, matching the four-policy arms' total interaction budget. The arm is a derived copy of the AM+rotate runner, so the clean single-factor contrast is AM+single against AM+rotate: both unfreeze the dispatcher, behaviour-clone it each iteration, and evaluate through it. What that dispatcher learns deserves an exact statement, because the label matters below: the cloning targets are the fixed schedule's own phase labels for the visited states (the driver's labelling loop assigns each state the scheduled module, despite a development-era docstring that claimed reward-based labelling, corrected in place in the artifact), so the ``learned dispatcher'' is a state-conditioned classifier distilled from the clock, not a reward-seeking dispatcher. Both arms share this, and both use the same \texttt{HybridLearner} with the same modality flags, the same governance configuration, and the same evaluation function, so success criterion, episode count, and reward accounting are identical. In both, reinforcement learning is the only modality that executes (\S\ref{sec:learning}). The single structural difference is that the module registry holds one whole-task module instead of four, so the schedule has nothing to rotate between and one policy drives the entire episode. The contrast against Ours additionally varies whether the dispatcher is frozen and whether evaluation routes through it, so we read that column as a reference point rather than as a controlled comparison. AM+single runs four seeds for ten iterations; each rotating arm has four seeds, the original two (7, 42) run for twenty iterations and the completion two (123, 2024) for eight, all cut to their first eight. All arms are compared over the first eight iterations, where cumulative \emph{training} interaction matches exactly; auxiliary episodes (five planner-labelling rollouts per iteration in the rotating and single arms alike, and per-module gate evaluations, three episodes each, so twelve per iteration for the rotating arms against three) are outside that budget and favour the rotating arms.

\noindent The factors this control does \emph{not} equalise are, on our reading, ones that favour the rotating arm, which is worth stating because it bears on the direction of residual confounds; we note the reading is contestable, since fewer parameters or earlier replay eviction can also act as sample-efficiency advantages. One whole-task module carries $107$k policy parameters against the four modules' $426$k, so the winning arm has four times less capacity. Its replay buffer begins evicting history from the sixth iteration while the rotating arm holds four buffers of $10^6$ transitions that never fill within eight iterations. The version gate is correspondingly stricter on the single module, since one gate decision governs the whole behaviour rather than one quarter of it. Cumulative training interaction over the compared window is equal by construction at $1.6$M SAC environment transitions per seed (auxiliary evaluation and labelling episodes differ and favour the rotating arms, as stated above); the rotating arms were run for longer in total, but those later iterations cannot enter a peak taken over the first eight. The unmatched quantities we identify do not obviously favour the arm that wins, though the direction of each is contestable.

\begin{table}[!ht]
  \centering
  \footnotesize
  \caption{Budget-matched comparison on T6 over the first eight iterations, so that every arm has consumed the same cumulative interaction budget. The first three rows train whole-task policies and differ in how many; the first spends its whole budget on one, the next two spread it over four that take turns on a schedule. The last two spend $20{,}000$ steps per iteration, their published default, and are shown for continuity. ``Past align'' is the fraction of evaluation episodes reaching the alignment step of the ladder of \Cref{sec:exp-mechanism} or beyond, measured over these same first eight iterations. These cells pool episodes over seeds and conceal seed spread, so we give the per-seed values (seed order 7, 42, 123, 2024): AM+single $0.871, 0.908, 0.342, 0.838$; Ours $0.683, 0.683, 0.558, 0.433$; AM+rotate $0.458, 0.392, 0.637, 0.463$; Agent Modification $0.058, 0.021$; SkiMo $0.067, 0.037$. AM+rotate's peak-success mean conceals one escaping seed, per-seed $3.3, 6.7, 3.3, 43.3$\%. No inference is drawn from these fractions.}
  \label{tab:nophase}
  \begin{tabular*}{\linewidth}{@{\extracolsep{\fill}}llccc@{}}
    \toprule
    Arm & Modules & Steps/iter & Peak success & Past align \\
    \midrule
    AM+single    & 1 whole-task            & $1\times200$k & $95.0 \pm 6.4$\% & 0.740 \\
    Ours         & 4 whole-task, rotated   & $4\times50$k  & $3.3 \pm 0.0$\% & 0.589 \\
    AM+rotate    & 4 whole-task, rotated   & $4\times50$k  & $14.2 \pm 19.5$\% & 0.488 \\
    \midrule
    Agent Modification & 1 monolithic & $1\times20$k & $0.0$\% & 0.040 \\
    SkiMo              & skill prior  & $1\times20$k & $0.0$\% & 0.052 \\
    \bottomrule
  \end{tabular*}
\end{table}

\begin{figure}[!t]
  \centering
  \includegraphics[width=\linewidth]{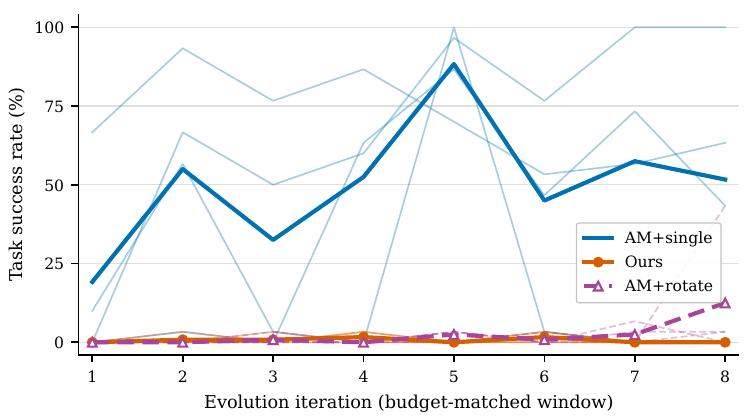}
  \caption{Per-iteration success over the budget-matched window. Thin lines are seeds, thick lines seed means. The single whole-task policy (AM+single, four seeds) climbs to peaks of $86.7$--$100\%$ within eight iterations; of the eight rotating-arm curves (four seeds per arm), seven stay at or below $6.7\%$ throughout, and the one exception is the learned dispatcher's escaping seed at $43.3\%$ (\Cref{sec:exp-nophase}). Rendered by \texttt{figures/render\_fig\_nophase\_curves.py} from the released per-seed result files.}
  \label{fig:nophase-curves}
\end{figure}

\noindent\textbf{Result.} The rotation does not pay for itself, and the margin is not subtle. At matched budget the single policy reaches $95.0 \pm 6.4$\% peak success, with per-seed values of $100$, $93.3$, $100$ and $86.7\%$ (seed order $7, 42, 123, 2024$), while the fixed schedule reaches $3.3$\% on all four of its seeds and the learned dispatcher $3.3$--$6.7$\% on three of four, with one escaping seed at $43.3$\% (\Cref{tab:nophase,fig:nophase-curves}). Peak reward moves the same way, $454$ to $471$ against $318$ to $395$ across the eight rotating seeds ($318$--$338$ for the fixed schedule, $341$--$395$ for the learned dispatcher), so this is not an artifact of the binary success criterion that makes T6 awkward elsewhere in this paper.

\noindent We hold this contrast to the same standard as the others in this paper. The rotating arms were first run at two seeds, where an exact Mann--Whitney test cannot return a two-sided $p$ below $2/15{=}0.133$; the completed design adds seeds 123 and 2024 for both rotating arms, run on the attested cloud stack and released with the rest. The completion holds: every single-policy seed sits above every rotating seed on retention, and the per-arm exact tests each return $p{=}2/70{=}0.029$ with $\delta{=}1.00$, their design floor and below $\alpha{=}0.05$. Corrected over the two per-arm contrasts the Holm-adjusted value is $0.057$, and we report that rather than hide it. Because the two rotating arms share the same four environment seeds, they cannot be pooled into eight independent observations; a seed-blocked combination that respects the sharing, taking each seed's better rotating retention as one observation, is also completely separated from the single policy and returns the same $p{=}0.029$, and we label it post hoc. The refutation rests on all of this together: the size and direction of the gap, its agreement with the retention mechanism below, the reading of the code, and a completed design whose exact tests reach their floor.

\noindent\textbf{Why it loses, which is not why we expected.} The rotating arms do not fail by never getting there. Over the budget-matched window of eight iterations, ours has the \emph{lowest} fraction of episodes that achieve nothing ($0.009$ pooled over four seeds) and drives the peg fully into the hole at some point in $28.0$\% of episodes, against $0.6$\% for monolithic Agent Modification. What it does not do is still be there at step 500. Writing retention for $P_4/(P_3{+}P_4)$, the fraction of episodes that reach the full success geometry and still hold it when the episode ends, and taking the seed rather than the episode as the unit of analysis, the single whole-task policy retains $0.813$, $0.817$, $0.788$ and $0.519$ across its four seeds, while the fixed schedule retains $0.019$, $0.034$, $0.021$ and $0.018$ and the learned dispatcher $0.043$, $0.067$, $0.040$ and $0.326$ (all in seed order $7$, $42$, $123$, $2024$) (\Cref{tab:retention}). The separation is complete at four seeds per arm, the weakest single-policy seed is $1.6$ times the strongest rotating seed and $15$ times the strongest fixed-schedule seed, and the per-arm exact tests reach significance at their completed design ($p{=}0.029$ each; the paragraph above gives the corrected and pooled forms). Retention describes what goes wrong better than peak success does, and at this design it also tests it. Retention for the two published baselines cannot be read at matched budget at all, since across both seeds they reach the success geometry in three and in eight episodes respectively. The environment's own success flag, read per step, tells the same story, though we note it is a second view of the same episodes rather than an independent instrument, because the ladder's top rungs are defined by the same geometric conditions the flag checks. Over its full twenty-iteration run on seed~7 the fixed-schedule arm reaches a state the environment scores as success and then loses it by the last step in $45.0$\% of episodes while holding it in $1.5$\%; this is not a warm-up effect, since over the final five iterations of that seed it loses it in $52.7$\% and holds it in $1.3$\%. The single policy, over its matched eight-iteration window on the same seed, loses it in $15.8$\% and holds it in $68.8$\%. The two figures come from different windows, and the direction is the same at either. A candidate mechanism is visible in the schedule rather than in the policies. The rotation has no feedback: window~7, steps $[434, 500)$, belongs to the fourth module in its second cycle regardless of what the robot has achieved, so whichever module drives the peg home is relieved of control at the next window boundary, at most one 62-step slice later (the final slice absorbs the horizon remainder and runs 66 steps), and hands a solved configuration to a policy whose own whole-task solution need not preserve it, while success is scored only at step~500. Put in the vocabulary of \Cref{sec:related}, the arrangement is an options architecture~\citep{sutton1999options} with the termination function replaced by a clock. Options are defined with a $\beta$ that can decline to return control, skill-chaining methods go to some length to make one skill terminate where the next can begin~\citep{konidaris2009chaining,bagaria2020dsc}, and the options literature prices switching itself as a cost~\citep{harb2018deliberation}. Our schedule has no such component, so it cannot express the one condition that matters here, that a module holding the peg in the hole should keep the arm. The failure is therefore predicted by the framework the configuration resembles rather than being a surprise: we omitted the component that framework specifies, and the system failed in the place that component exists to cover. That is an argument for reporting it. A configuration that drops $\beta$ is easy to build without noticing, because a fixed schedule is the simplest thing that dispatches modules, and this paper is a record of one team building it, describing it as a phase decomposition, and measuring what the omission cost. The account so far is observational: it comes from logs that were not collected to test it. We therefore ran the intervention it implies.

\noindent\textbf{The intervention.} We re-evaluated the fixed-schedule arm's own trained modules under two dispatch rules that differ in one line. The \emph{clock} arm is the released schedule. The \emph{hold} arm is identical until the success geometry is first satisfied; from that step the module then in control keeps it to the end of the episode, which is the termination condition the clock replaced. No training happens in either arm, the modules are the same checkpoints, and the intervention is applied only to the fixed-schedule arm because that is where the diagnosed mechanism sits. Every one of the four seeds improves (\Cref{tab:handoff}): retention rises from a seed mean of $0.112$ to $0.571$ and success from $3.3\%$ to $26.7\%$. Those two are not independent confirmations, and we say so as we did for the environment flag earlier: both count the same episodes in their numerator, $P_4$, and differ only in the denominator, the episodes that reached the geometry against all thirty. The hold engages in every episode that reaches the geometry, which is what the rule specifies rather than something measured, at a mean step of $98$ to $304$ depending on the seed, well before the horizon. Two properties of the comparison need stating before the numbers are read. This evaluation runs each arm's thirty episodes on the final checkpoint of the run, whereas \Cref{tab:retention} pools the $240$ episodes of a seed's first eight iterations, so the clock column here is a different estimate of the same quantity, not the same number: seed $42$ reads $0.375$ on the final checkpoint against $0.034$ pooled over its window. Pooling the clock arm's four seeds gives $4/58{=}0.069$. That is not the matched-window pooled figure, which is $6/269{=}0.022$; it happens to sit near the full-run pooled $38/561{=}0.068$, a window this paper reports but declines to read as a property of the arm (\Cref{tab:retention}). The per-seed mean of $0.112$ is carried by the single $3/8$ cell, so we take the pooled $0.069$ as this table's estimate of record and report the mean beside it. And the intervention covers the fixed-schedule arm only: no module weights survive for the learned-dispatcher runs, so its four cells failed on missing checkpoints and we could not evaluate them; whether those runs ever wrote weights is not something the surviving record settles. The direction is unanimous at four seeds out of four, which an exact two-sided sign test cannot push below $p{=}2/2^4{=}0.125$; that floor is a property of the design, not of the effect, and we report the effect size rather than claim significance. Two limits belong with the result. The arms are paired at the level of seed and checkpoint rather than episode by episode, because the simulator's placement sampler is not seeded (\Cref{sec:provenance}), so each arm draws its own initial states. And holding control is not a dispatcher anyone would ship: it is the minimal edit that tests whether the unconditional handoff is what costs the reached goals. Within those limits the mechanism is no longer only consistent with the data; removing the handoff removes most of the loss.

\begin{table}[!ht]
  \centering
  \footnotesize
  \caption{State-conditioned handoff intervention on the fixed-schedule arm, evaluation only, 30 episodes per arm per seed on that seed's final checkpoint. \emph{Clock} is the released schedule; \emph{hold} freezes control in whichever module is acting when the success geometry is first reached. Retention is $P_4/(P_3{+}P_4)$ with the same ladder as \Cref{tab:retention}, but over these 30 final-checkpoint episodes rather than that table's 240 pooled window episodes, so the clock column is a second estimate of that arm's retention and not the same number; ``reached'' gives each cell's denominator. Retention and success share the numerator $P_4$. Produced by \texttt{handoff\_runner.py}; the driver, the per-seed outputs, and the module weights it evaluates are all in the artifact, so this table can be regenerated from a clean checkout.}
  \label{tab:handoff}
  \begin{tabular*}{\linewidth}{@{\extracolsep{\fill}}lcccccc@{}}
    \toprule
    & \multicolumn{3}{c}{Clock} & \multicolumn{3}{c}{Hold} \\
    \cmidrule(lr){2-4}\cmidrule(lr){5-7}
    Seed & Reached & Retention & Success & Reached & Retention & Success \\
    \midrule
    $7$    & $14$ & $0.071$ & $3.3$\%  & $9$  & $0.556$ & $16.7$\% \\
    $42$   & $8$  & $0.375$ & $10.0$\% & $13$ & $0.462$ & $20.0$\% \\
    $123$  & $9$  & $0.000$ & $0.0$\%  & $9$  & $0.667$ & $20.0$\% \\
    $2024$ & $27$ & $0.000$ & $0.0$\%  & $25$ & $0.600$ & $50.0$\% \\
    \midrule
    Seed mean & --- & $0.112$ & $3.3$\% & --- & $0.571$ & $26.7$\% \\
    Pooled & $58$ & $0.069$ & --- & $56$ & $0.571$ & --- \\
    \bottomrule
  \end{tabular*}
\end{table}

\noindent\textbf{The multi-module advantage fails, and so does the description it rested on.} Two things go. The performance separation in \Cref{tab:method-peak-success} is an artifact of unequal compute, and at matched budget it reverses. And the thing that separation was attributed to did not exist in the code: there was no phase-scoped decomposition, only four whole-task policies rotated on a fixed schedule, so the system's stored description of its own implementation was wrong rather than merely overstated. What we refute is the arrangement we actually built, rotating control among independently trained whole-task policies on a fixed schedule. We refute it on one task, with one learner, at one budget. The mechanism above is what makes that narrow result worth reporting: the arrangement reaches the goal in a third of its episodes and then fails to hold it, far more often than the arm that wins, and the loss coincides with a handoff that a line of scheduling code makes unconditional.

\noindent\textbf{Two rescues, and how far each goes.} The first is that the fixed schedule starved the modules, so what failed was the dispatcher rather than the structure. The AM+rotate arm addresses part of this. A learned dispatcher over the same four modules lands at $3.3$--$6.7$\% peak success on three of its four seeds, indistinguishable from the fixed schedule, whose four seeds all sit at $3.3$\%; its fourth seed is the exception this completion surfaced, reaching $43.3$\% peak success and a retention of $0.326$. What the four-seed design shows is window-dependent. Over the matched window the two dispatchers separate completely at four seeds each (every learned-dispatcher seed above every fixed-schedule seed, exact $p{=}0.029$), driven partly by the escaping seed; over full runs the per-seed values still interleave (exact $p{=}0.114$), so we read the matched-window separation as suggestive and not as an established ordering. Both arms sit below every single-policy seed on every quantity. What the arm does bound is this dispatch policy: a classifier distilled from the clock reproduces the clock's handoffs on three of four seeds, and its fourth seed shows the difference between a clock and a state-conditioned imitation of one, holding the goal in $0.326$ of reached episodes over this window, where no fixed-schedule seed exceeds $0.034$ over the same window. A dispatcher with state freedom sometimes declines the destructive handoff; the clock never can. A genuinely reward-seeking or value-based dispatcher is not tested by this arm, so nothing here bears on learned dispatch in general. The handoff structure itself is ablated only in the minimal form reported above, where control freezes once the goal geometry is reached; a dispatcher that decides when to hand over, rather than simply never handing over after success, remains untested. The second rescue is that splitting a fixed budget four ways starves each module of data. This is true, and it is not a defence: the only way to give each module the data it wants is to grant the rotating arm four times the total compute, which is the confound this control exists to remove. The claim fails at matched budget and is untestable at unmatched budget.

\noindent\textbf{What this does not establish.} We refuted one arrangement, not modular structure, and certainly not phase decomposition, which this system never implemented and this study therefore never tested. Both rotating arms share the same four whole-task policies, the same fixed window boundaries, the same interfaces, and the same uniform allocation of $50{,}000$ steps per module; varying the dispatcher over that design ablates the dispatch policy alone. Modules trained on sub-problems with phase-scoped resets, rewards, and terminations, an allocation that gives harder segments more data, a dispatcher that decides from general state when to hand over rather than only suppressing handover once the success geometry is reached, or modules sharing a backbone might all behave differently, and none of these is tested here beyond the minimal hold intervention reported above. The evidence is also from one task and one learning algorithm. What we claim is narrow and, we think, still worth reporting: on T6, at matched interaction budget, under SAC, every seed of the single-policy arm outperformed every seed of both rotating arms on peak success, peak reward, and retention, and at the completed four-seed design the per-arm exact tests certify the retention ordering ($p{=}0.029$ each; corrected and pooled forms above). The loss is a loss of an already-reached goal state. The fixed schedule cannot avoid it on any seed; the learned dispatcher escaped it on one seed of four, so the dispatch policy modulates the failure without removing the deficit.

\subsection{Where Episodes Stall, and What They Fail to Hold (H4)}
\label{sec:exp-mechanism}

The comparisons so far order the arms by peak success without saying what the ordering is made of. We instrument the evaluation to see how far episodes get, and what becomes of the ones that get all the way.

\noindent\textbf{Setup.} T6 requires two arms to bring a peg and a plate together and insert one into the other. The simulator exposes the geometry this depends on as a projection $t$, a distance $d$, and an alignment cosine, and defines success as $d < 0.06$ with $-0.12 \le t \le 0.14$ and $\cos > 0.95$. We use these same quantities to place each evaluation episode on a four-step ladder: \emph{approach} ($P_1$) when $d$ falls below $0.12$, \emph{align} ($P_2$) when the alignment criterion is met while approached, \emph{insertion} ($P_3$) when the full success geometry holds at some step, and \emph{success} ($P_4$) when the episode terminates successfully; $P_0$ is an episode that reaches none of these. Two derived quantities matter below. Progress is the fraction of episodes at $P_2$ or beyond. \emph{Retention} is $P_4/(P_3{+}P_4)$, the fraction of episodes that get the peg fully into the hole and are still holding it when the episode ends. We ran all four methods with this instrumentation for 20 iterations on seeds $\{42,7\}$, and additionally recorded, once per iteration, each method's decision output on a fixed set of 100 probe states drawn independently of any method. The single-policy control of \Cref{sec:exp-nophase} carries the same instrumentation over four seeds and ten iterations. Retention is computed per seed and summarised across seeds, for the reason given below \Cref{tab:retention}; the three arms trained at $200{,}000$ environment steps per iteration consume the same budget each iteration, so cutting them to a common iteration count matches cumulative interaction exactly, and the budget-matched window is the first eight iterations; the two published baselines run at their $20{,}000$ default throughout and are never budget-matched to the other three.

\begin{table}[!ht]
  \centering
  \footnotesize
  \caption{Fraction of evaluation episodes reaching the align step or beyond on T6, and mean per-iteration drift of the decision layer and of the emitted actions on the shared probe set. Two seeds, 20 iterations, 30 episodes per iteration.}
  \label{tab:mechanism}
  \begin{tabular*}{\linewidth}{@{\extracolsep{\fill}}lccc@{}}
    \toprule
    Method & Reached align or beyond & Decision drift & Action drift \\
    \midrule
    Ours (4 rotating, frozen)   & 0.744 & 0.000 & 0.65 \\
    AM+rotate (4 rotating)      & 0.547 & 0.766 & 0.73 \\
    Agent Modification          & 0.146 & n/a   & 0.31 \\
    SkiMo                       & 0.050 & n/a   & 0.32 \\
    \bottomrule
  \end{tabular*}
\end{table}

\begin{table}[!tp]
  \centering
  \footnotesize
  \setlength{\tabcolsep}{3pt}
  \caption{Retention on T6 with the seed as the unit of analysis. Retention is $P_4/(P_3{+}P_4)$, the fraction of evaluation episodes that reach the full success geometry and still hold it at the final step. AM+single trains one whole-task policy; AM+rotate and Ours rotate four of them, under a learned dispatcher and under a fixed schedule respectively. ``Reached'' is the $P_3{+}P_4$ episode count that forms that seed's denominator, and seeds appear in the order $7$, $42$, $123$, $2024$. Each seed contributes 30 episodes per iteration, so 240 episodes over the matched window; over the full run, 300 for AM+single, 600 for the rotating arms' original seeds (7, 42), and 240 for their completion seeds (123, 2024), which ran the matched window only, on the attested cloud stack. The upper block cuts every arm to its first eight iterations, where cumulative training interaction is $1.6$M SAC environment transitions per seed for the three arms trained at $200{,}000$ steps per iteration; Agent Modification and SkiMo run at their published $20{,}000$ default over those same eight iterations and are never budget-matched to the other three. The lower block uses each arm's full run (ten iterations for AM+single, twenty for the rotating arms' original seeds, eight for their completion seeds), which is not budget-matched and is shown so the matched-window reading can be checked against the arms' full histories. Episodes are clustered inside seeds, so the summary is the seed mean and the seed range and no interval here is computed from pooled episodes. At matched budget the two published baselines reach the geometry in three and eight episodes in total, so their retention is not estimable there. Produced by \texttt{analysis/retention\_seed\_level.py} in the artifact, which regenerates this table from the released per-seed files.}
  \label{tab:retention}
  \begin{tabular*}{\linewidth}{@{\extracolsep{\fill}}lllcc@{}}
    \toprule
    Arm & Reached, per seed & Retention, per seed & Mean & Range \\
    \midrule
    \multicolumn{5}{@{}l}{\emph{Budget-matched: first eight iterations, $1.6$M training transitions per seed}}\\
    AM+single          & 203, 208, 66, 183 & 0.813, 0.817, 0.788, 0.519 & $0.734$ & 0.519--0.817 \\
    AM+rotate          & 46, 45, 100, 43   & 0.043, 0.067, 0.040, 0.326 & 0.119 & 0.040--0.326 \\
    Ours               & 107, 59, 48, 55   & 0.019, 0.034, 0.021, 0.018 & 0.023 & 0.018--0.034 \\
    Agent Modification & 2, 1              & 0.000, 0.000               & \multicolumn{2}{c}{not estimable} \\
    SkiMo              & 6, 2              & 0.000, 0.000               & \multicolumn{2}{c}{not estimable} \\
    \midrule
    \multicolumn{5}{@{}l}{\emph{Full run, not budget-matched: ten iterations for AM+single against twenty}}\\
    AM+single          & 263, 258, 68, 243 & 0.829, 0.829, 0.779, 0.481 & 0.730 & 0.481--0.829 \\
    AM+rotate          & 135, 216, 100, 43 & 0.141, 0.282, 0.040, 0.326 & 0.197 & 0.040--0.326 \\
    Ours               & 279, 179, 48, 55  & 0.032, 0.151, 0.021, 0.018 & 0.056 & 0.018--0.151 \\
    Agent Modification & 129, 3            & 0.705, 0.333               & 0.519 & 0.333--0.705 \\
    SkiMo              & 34, 2             & 0.353, 0.000               & 0.176 & 0.000--0.353 \\
    \bottomrule
  \end{tabular*}
\end{table}

\noindent\textbf{Where episodes stall.} Read within a fixed budget, progress along the ladder tracks interaction rather than structure. At $20{,}000$ steps per iteration the two monolithic baselines stall early and get worse as training proceeds: across the 20 instrumented iterations the fraction of episodes achieving nothing at all moves from an early-band mean of $0.29$ to a late-band mean of $0.87$ for SkiMo, and from $0.28$ to $0.68$ for Agent Modification (non-monotonically, peaking at $0.79$ mid-run), so these policies do not fail at insertion but lose the approach they had. The rotating configurations, at $200{,}000$ steps per iteration, hold that fraction far lower over the same 20 iterations, at $0.008$ under the fixed schedule and $0.070$ under the learned dispatcher, and carry $0.744$ and $0.547$ of episodes past alignment (\Cref{tab:mechanism}). Over the shorter eight-iteration window on which the budget-matched control is compared, the two rotating arms sit at $0.009$ and $0.077$ on the same fraction (pooled over their four seeds each) and carry $0.589$ and $0.488$ past alignment (\Cref{tab:nophase}). It would be easy to read this as the module structure doing the work, and the stored documentation records exactly that reading. The control in \Cref{sec:exp-nophase} shows the reading was wrong: one policy at the same $200{,}000$ steps carries $0.740$ past alignment over that same eight-iteration window, more than either rotating arm reaches over it. Progress separates budgets, and about structure it says only that the starved baselines fail at approach rather than at insertion.

\begin{figure}[!t]
  \centering
  \includegraphics[width=\linewidth]{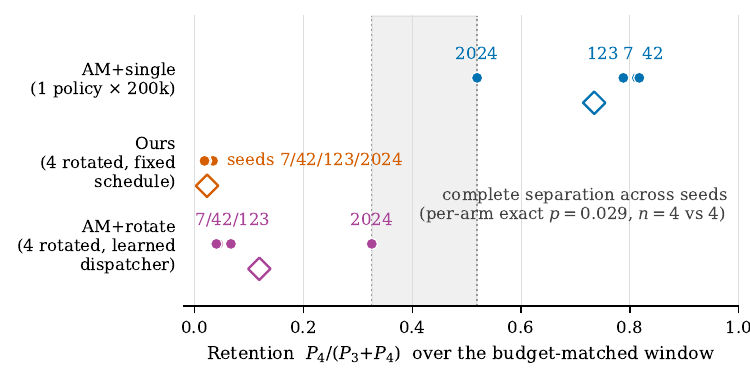}
  \caption{Seed-level retention over the budget-matched window. Circles are individual seeds (labelled), diamonds are seed means; the shaded band is the empty interval between the strongest rotating seed ($0.326$) and the weakest single-policy seed ($0.519$). The separation is complete at four seeds per arm, and each per-arm exact test returns its design floor of $p{=}0.029$. Rendered by \texttt{figures/render\_fig\_retention.py} from \texttt{retention\_seed\_level.json}.}
  \label{fig:retention}
\end{figure}

\noindent\textbf{What distinguishes the arms is retention.} \Cref{tab:retention} and \Cref{fig:retention} report the quantity that does discriminate, and progress and retention order the arms differently. The fixed-schedule rotating arm is the best arm in the study at leaving the starting line, with the lowest $P_0$ ($0.008$ over its 20 instrumented iterations) and the highest rate of leaving the initial step ($0.992$), and it reaches the full insertion geometry in $38.2$\% of episodes against $11.0$\% for monolithic Agent Modification over the same 20 iterations. It is the worst arm in the study at staying there. Over the budget-matched window its four seeds hold the geometry at the final step on $0.018$ to $0.034$ of the episodes that reached it, against $0.519$ to $0.817$ for the single whole-task policy; the learned dispatcher gives $0.040$ to $0.326$, its highest seed being the escape discussed in \Cref{sec:exp-nophase}. Every seed of the single-policy arm sits above every seed of both rotating arms ($p{=}0.029$ per arm at the completed design). Against the single policy the fixed-schedule arm is behind on both counts, and by very different margins: over the matched window it reaches the geometry in $28.0$\% of episodes against $68.8$\%, a factor of $2.5$, and holds what it reaches at roughly one thirtieth the rate.

\noindent The two published baselines drop out of this comparison rather than joining it. At matched budget they reach the success geometry in three episodes (Agent Modification) and eight (SkiMo) across both seeds, and hold it in none, so retention has no denominator to speak of. Over their full twenty iterations Agent Modification reads $0.705$ on seed~7 and $0.333$ on seed~42, but the first of those rests on 129 episodes and the second on three, so it is a one-seed quantity and we do not use it to argue that durability is a property of monolithic policies.

\noindent\textbf{Why the table reports seeds rather than pooled intervals.} Evaluation episodes are not independent draws. They are clustered inside seeds, and a pooled binomial prices only the within-episode noise, so an interval computed from it is far narrower than the design warrants. A chi-square test of equal retention across seeds within an arm rejects for AM+single and AM+rotate over the matched window ($\chi^2{=}57.6$ and $31.0$, $\mathrm{df}{=}3$, $p{<}10^{-6}$) and for all three budget-matched arms over their full runs; the one exception is Ours over the matched window ($\chi^2{=}0.5$, $p{=}0.92$), whose four seeds sit within $0.018$--$0.034$. Resampling seeds rather than episodes puts AM+single at $[0.590, 0.815]$ over the matched window; the released script reports cluster-bootstrap intervals over the seeds of each arm, four for the three budget-matched arms and two for the published baselines, and \Cref{tab:retention} reports the seed range alongside them.

\noindent\textbf{What retention therefore supports.} A contrast with a large effect, a cause we can point to in the source, and, at the completed four-seed design, a significant seed-level exact test ($p{=}0.029$ for AM+single against each rotating arm, with the Holm-corrected and pooled forms stated in \Cref{sec:exp-nophase}). Moving the claim from peak success to retention buys a decomposition of the outcome into reaching and holding, and a failure mode that is identifiable in the scheduling code, namely the unconditional handoff described in \Cref{sec:exp-nophase}. The reading is corroborated by the environment's own success flag, which fires at some step in $46.5$\% of the fixed-schedule arm's episodes on seed~7 and survives to the last step in $1.5$\%.

\noindent\textbf{What the drift numbers say.} Two of these columns cut against a tempting explanation. The decision layer of our method registers exactly zero drift on the probe set across every iteration of both seeds. This number is weaker than it looks and we say so: in the evaluated configuration the learned planner is held with gradients disabled and the fixed time-slice schedule is what routes actions, so the planner is not on the evaluation path and a drift of zero is what the construction dictates. What the measurement adds is an audit, that nothing in the training loop writes to those parameters by accident, which is worth checking and is not the same as an empirical result. The comparison that does carry information is the same architecture with the layer unfrozen, which registers $0.766$ (per-seed $0.767$ and $0.766$) and therefore shows the probe is capable of detecting drift when drift exists. Yet at the level of emitted actions our method is the \emph{most} variable of the four, at $0.65$ against $0.31$ for Agent Modification, whose action drift settles to about $0.14$ as its policy converges (non-monotonically; eight of its thirty-eight adjacent steps increase). The methods whose actions stop changing are the ones stuck at approach, and the arm whose actions change most gets furthest along the ladder while retaining least. Instability of the emitted behaviour is therefore not what distinguishes these arms either; the retention contrast is, and \Cref{sec:exp-nophase} points to the schedule rather than to any policy's parameters.

\noindent\textbf{What holds across every cut of this data.} The retention gap between the single policy and the rotating arms holds in every seed of every arm, over the matched window and over the full runs alike, while the peak-success column underneath swings by a factor of three within a single method (\Cref{tab:method-peak-success}). The gap between the two rotating arms is window-dependent: complete over the matched window at four seeds each, interleaved over their full runs, where the fixed schedule's better seed ($0.151$) sits above the learned dispatcher's seed 7 ($0.141$), so we do not treat the two arms as ordered. The fixed schedule's seed-42 rise from $0.034$ to $0.151$ between the two blocks comes entirely from iterations 9--20, an improvement its seed 7 shows only weakly over the same span ($0.019$ to $0.032$), so it reads as one stream's late drift rather than a property of the arm. We place the mechanism claim on the ladder because the ladder splits the outcome into reaching and holding and the holding failure is identifiable in the scheduling code; the completed four-seed tests of \Cref{sec:exp-nophase} clear their design floor, and we cite them there rather than re-litigate them here. Read plainly, the behaviour of the peak-success column is a limitation of the benchmark rather than a property of any method: a task whose binary criterion fires this rarely cannot rank methods at the seed counts a study like this can afford. We take the practical consequence seriously in \Cref{sec:discussion}.

\section{Discussion}
\label{sec:discussion}

\looseness=-1 Three findings carry the study, and each is told in full at its site; here we only weigh them. First, the architectural refutation (\Cref{sec:exp-nophase,sec:exp-mechanism}): the rotation loses at matched budget, its characteristic loss being a reached goal surrendered at a clock boundary, which an intervention restoring the termination condition largely removes, the failure the options framework's $\beta$ exists to prevent. The identity-invariance property (\Cref{eq:identity-invariance}) is untouched by this: it is a design-level benefit of a frozen decision layer (stability, auditability, rollback), it holds by construction, and the matched pair of \S\ref{sec:exp2} shows it costs nothing resolvable at the attested seeds. What it was paired with on this task was the wrong structure. Second, the governance result (\Cref{sec:exp3,sec:exp-govstress}): what replaces the claim is about our measurement rather than about the layer, and its general form is stated once in \Cref{sec:exp3} and once below. Third, the lifecycle measurements (\Cref{sec:exp-lifecycle}): a promotion gate whose interception is near-guaranteed by calibration but whose clean-candidate rejection cost is a genuine measurement, and a dip detector whose misses cluster in one seed with a severity sweep locating its silent region. A natural design question is whether full freezing is the right choice; \emph{partial} adaptation (e.g.\ updating ECM-selection heuristics within $\pi_{\mathcal{A}}$ while keeping identity memory $\mathcal{M}_{\mathcal{A}}^{\text{id}}$ fixed) occupies a continuum between our fully frozen agent and fully mutable baselines, and we adopt full freezing for the strongest stability guarantee.

\noindent\textbf{What we are and are not competing on.} Two systems released during this work, PhyAgentOS~\citep{liu2026phyagentos} and CaP-X~\citep{fu2026capx}, target evolving embodied agents at a scale this study does not approach: a full operating system evaluated on real robots in one case, a benchmark spanning twelve frontier models and hundreds of tasks in the other. We are not offering a competing capability claim, and a reader comparing peak task performance should prefer their numbers to ours.

The distinction that makes our question a separate one is what each system evolves. PhyAgentOS consolidates verified outcomes into an epistemic memory and states explicitly that this closes the loop without retraining; CaP-X improves programs that an agent writes, which can be read, diffed, and reviewed. Both artifacts are legible: a lesson can be inspected, a program can be code-reviewed, and either can be reverted by ordinary means. A recent survey organises robot learning along this same axis, learned weights against written code, and finds that only the code sense of a skill improves without gradient updates~\citep{jena2026weights}. What we evolve is a set of learned parameters, and that artifact is legible in none of those ways. A retrained module ships no changelog, cannot be exhaustively unit-tested in the ordinary deterministic sense, and can regress without any symptom until it is executing on a robot. Opaque parameters change the evidence an admission decision can rest on: a program can be reviewed before it ships, whereas a retrained module can be judged only by how it behaves on held-out episodes. That is why the question takes a statistical form in our setting, and we do not read the absence of version registries, promotion gates, or rollback paths in either system as an oversight. Among concurrent systems that do evolve learned policies the picture is mixed: AgenticRobotics gates promotion on recorded evidence~\citep{yu2026agenticrobotics}, while ENPIRE and REVOLVE retrain real-robot policies in closed loops without reporting an admission gate or a rollback path~\citep{xiao2026enpire,liu2026revolve}, and the gate-off arm of \Cref{sec:exp-lifecycle} measures what that absence costs.

This also bounds what we claim. Enforcement at the execution boundary is not ours to claim as novel; PhyAgentOS layers preflight checks, action bridges, and a target-side guard, and shielding was an execution-boundary construct before any of this work. The design intent was to pair such a boundary with a module lifecycle so that the envelope would be invariant across versions of what it governs, and we cannot report that pairing working, because our realisation of the boundary stops the robot. What we can report is the cost of each mechanism, which is the useful part. A promotion gate that admits nothing bad also rejects $22.5\%$ of good candidates; a dip detector with zero triggers on the clean iterations we observed also misses three of twelve deployed regressions, all on one seed; and an action-level shield scored on violations alone scores near-perfectly while completing no task at all. Systems papers are read by people who will have to live with those numbers, and a mechanism characterised down to where it stops working is more useful to them than another demonstration of an upper bound.

\noindent\textbf{What the peak instability costs the field.} The peak-instability negative result (\Cref{sec:exp2}) has a practical consequence worth drawing out. If peak binary success on this task is not stable across random streams at the seed counts we can afford, then it cannot serve as the endpoint of a comparison, and neither can any quantity derived from a single run of it. This is not a statement about our seeds being too few for significance; it is a statement that the quantity itself is unstable enough that a ranking read from it changes with the random stream alone. Work that needs a comparison on a long-horizon manipulation task should either adopt an endpoint with better stability evidence than single-run peaks, as retention has here (its separation holds at every seed and at both analysis windows, and the environment's own success flag agrees, though the instrumented runs behind it were collected once and the measurement has not itself been repeated), or budget for seed counts that the T6 variance actually requires, since three seeds per method, at the spread in \Cref{tab:method-peak-success}, do not establish a stable ranking; standard power analysis for reinforcement learning shows how to size that count for a given effect~\citep{colas2018seeds}. We report this because the literature this paper sits in routinely ranks methods on exactly this endpoint at exactly these seed counts, a practice the deep-RL evaluation literature has warned against for years~\citep{henderson2018matters}, with interval-based aggregates prescribed precisely because point estimates at these run counts mislead~\citep{agarwal2021precipice}; the seed spread here is a systems-side instance of the warned failure. A concurrent study of automated robot-policy improvement reports a related failure from the selection side: on the one lineage it measured, the checkpoint that selection tasks ranked highest was not the one held-out tasks ranked highest, so selecting on a measured score misled~\citep{yu2026agenticrobotics}.

\noindent\textbf{The metric that omits the system.} The governance result is another instance in this study of a described behaviour turning out not to be the implemented one, and it differs from the others in an instructive way. Phase decomposition, imitation learning, and code synthesis were absent, and no measurement could have shown them working because none of them ran. The shield ran. It was measured, the measurement was clean, and the measurement was of the wrong thing. Every quantity we collected about it (violations prevented, actions blocked, microseconds spent) improves monotonically as the shield becomes more aggressive, and the limiting case of a shield that rejects everything is the global optimum of that criterion. Nothing in the acceptance criterion was about the robot still doing its job. We take two rules from this and offer them as the transferable part of the finding. A mechanism claim needs paired evidence, that the mechanism fired and that the system it acts on still functions, and a report of one without the other should be read as incomplete rather than as good news. And a safety envelope should be validated against a known-good trajectory before it is validated against attacks, since an envelope that admits no trajectory at all will pass every adversarial test that exists.

\emph{Limitations.}
(i)~The refutation is of one arrangement on one task with one learner, and it is not evidence about modular structure in general. Nothing here tests modules trained on sub-problems, since the system never implemented them. The natural next step is to build an actual phase decomposition, modules whose training is scoped to a segment of the task by its own resets, rewards, and terminations, together with a dispatcher conditioned on general state rather than on a step counter, going beyond the minimal post-success hold tested here, and to compare that against a single whole-task policy at matched budget; validation on real hardware with richer governance specifications is the other open direction.
(ii)~The training-based comparison is concentrated on T6; the suite's shorter tasks enter only through the evaluation-only governance stress test (\Cref{sec:exp-govstress}), so this paper reports no training result on them.
(iii)~The five-method comparison rests on three seeds per method, with no hypothesis test attached; these observations do not establish a stable ranking; the budget-matched contrast this paper's architectural claim rests on runs four seeds per arm and reaches its design floor. Widening the method comparison to seed counts where tests bind, and adopting a standard lifelong-robot benchmark, are the natural next steps.
(iv)~The apparent multi-module advantage in \Cref{tab:method-peak-success} does not survive scrutiny on either of its two legs. The confound that matters is the interaction budget, since the multi-module arms spend ten times what the monolithic ones do, $200{,}000$ environment steps per iteration against $20{,}000$, and the control in \Cref{sec:exp-nophase} equalises it and reverses the conclusion. And there is no decomposition to credit: the modules are four whole-task policies with phase labels, and the label never enters training (\Cref{sec:agent}). What replaces the claim is narrower and has a mechanism: rotating control among independently trained whole-task policies on a fixed schedule loses goal states it has already reached, at a per-seed retention of $0.018$ to $0.034$ across four seeds against $0.519$ to $0.817$ for one policy holding control throughout, a loss the unconditional handoff accounts for: freezing control in whichever module reaches the goal raises retention on every seed (\Cref{sec:exp-nophase}). That contrast is large and, at the completed four-seed design, significant at its exact-test floor ($p{=}0.029$ per arm). What we can still say about freezing is that the frozen and unfrozen variants have equal mean peak success at the three attested seeds, which is a failure to separate rather than a demonstration of equivalence, and that the freeze is implemented as declared (\Cref{sec:identity}). The same failure of description reached the learning modalities, where imitation and synthesis were reported as active and neither ever executed (\S\ref{sec:learning}). We found both errors by reading the code, and no result had pointed at either, which is the part we would ask a reader to take from this: a component that is configured, logged, and cited in a paper can still never have run.
(v)~No safety claim survives for the governance layer, and the reason is broader than the two defects behind it. The keep-out zone can be moved outside the workspace and the end-effector position can be routed into the predicates, and we expect both fixes to take an afternoon, but a repaired shield would still be unevaluated: no measurement in this paper reports a task completed under governance, because no run in this paper completed one. Until a repaired layer is shown to hold the envelope while the six tasks still reach the success rates they reach without it, this work supports no claim about the cost of runtime governance. We also do not know how far the second defect reaches on its own, since the stress test that used real positions changed the predicate inputs and the outcome did not move, which tells us the zone placement is sufficient to paralyse the arm but not what the wiring error costs when the zone is correct. Separating the two requires a run we have not done.

\section{Conclusion}
\label{sec:conclusion}

\looseness=-1 We have presented a runtime architecture for robot systems whose skills keep changing after deployment, and an evaluation that refuted two of the four premises it was built on and inverted the expected reading of a third. Learned functionality lives in versioned modules with a promotion gate and a rollback path, and the decision layer that dispatches them is held fixed, so the provenance of any deployed behaviour is unambiguous and freezing it costs nothing resolvable at the seeds run here. Our contributions are: (i)~a lifecycle for learned skill modules covering versioning, gated promotion, and rollback, measured under fault injection: the gate kept all twelve injected regressions out of deployment, a rate its calibration nearly guarantees, at a $22.5\%$ clean-candidate rejection cost, and the dip detector caught nine of twelve deployed regressions with zero triggers on the clean iterations observed, its misses clustered in one seed (\Cref{sec:exp-lifecycle}); (ii)~a formal model casting Embodied Capability Modules (ECMs) as modular, versioned units of embodied functionality with a closed-loop evolution operator; (iii)~a negative result about how a runtime safety layer should be evaluated, which rules out any safety claim for the realisation evaluated here; and (iv)~an empirical study on T6 showing that the rotation the system actually implements does not pay for its budget, and that what it loses is retention of already-reached goals.

The three negative results are each told in full exactly once, and we only cite them here. A shield scored on violation counts alone looked near-perfect while discarding the policy's whole action on most steps and holding task success at zero; the transferable form is that a safety mechanism whose acceptance criterion omits task completion is optimised by stopping the system it protects (\Cref{sec:exp3,sec:exp-govstress}). A rotation of four whole-task policies, which the system's stored description calls a phase decomposition, loses at matched budget to one policy given the same interaction, its losses concentrated in episodes that reach the goal and then surrender control, the failure pattern an options termination function exists to prevent (\Cref{sec:exp-nophase,sec:exp-mechanism}). And single-run peak success swings by a factor of three across seeds within one method, so that endpoint cannot rank methods at these seed counts (\Cref{sec:exp2}). Open directions include repairing and then genuinely evaluating the governance layer against task completion, state-conditioned handoff in place of the clock, parameter-level continual-learning comparisons, standardised benchmarks for capability evolution, formal interface contracts across skill versions, and multi-agent extensions with shared skill registries evolving asynchronously.


\section*{CRediT authorship contribution statement}

\textbf{Xue Qin}: Conceptualization, Methodology, Software, Investigation, Formal analysis, Writing -- original draft. \textbf{Simin Luan}: Validation, Writing -- review \& editing. \textbf{Cong Yang}: Supervision, Writing -- review \& editing. \textbf{Zhijun Li}: Supervision, Writing -- review \& editing.

\section*{Declaration of competing interest}

The authors declare that they have no known competing financial interests or personal relationships that could have appeared to influence the work reported in this paper.

\section*{Data availability}

The replication artifact accompanying this submission contains the seven evaluation drivers, the per-seed result files behind every number the evaluation reports, checksummed checkpoints for the handoff and detector-severity evaluations, and the analysis and figure scripts, with dependencies pinned; the code line is hosted at \url{https://github.com/s20sc/capability-evolution}. The Artifact availability paragraph in \Cref{sec:exp-setup} states what the artifact contains, and the governance-stress setup (\Cref{sec:exp-govstress}) states which replay inputs were not preserved.

\section*{Funding}

This research did not receive any specific grant from funding agencies in the public, commercial, or not-for-profit sectors.


\bibliographystyle{elsarticle-harv}
\bibliography{references}

\end{document}